\documentclass[sigconf]{acmart}
\copyrightyear{2026}
\acmYear{2026}
\setcopyright{cc}
\setcctype{by}
\acmConference[KDD '26]{Proceedings of the 32nd ACM SIGKDD Conference on Knowledge Discovery and Data Mining V.1}{August 09--13, 2026}{Jeju Island, Republic of Korea}
\acmBooktitle{Proceedings of the 32nd ACM SIGKDD Conference on Knowledge Discovery and Data Mining V.1 (KDD '26), August 09--13, 2026, Jeju Island, Republic of Korea}
\acmPrice{}
\acmDOI{10.1145/3770854.3780259}
\acmISBN{979-8-4007-2258-5/2026/08}
\newcommand{\QQQ}{RUQuant }

\usepackage{amssymb,amsmath}
\usepackage[utf8]{inputenc} 
\usepackage[T1]{fontenc}    
\usepackage{hyperref}       
\usepackage{url}            
\usepackage{multicol}      
\usepackage{subfigure} 
\usepackage{graphicx}      
\usepackage{booktabs}       
\usepackage{amsfonts}       
\usepackage{nicefrac}       
\usepackage{microtype}      
\usepackage{xcolor}         
\usepackage{color, colortbl}
\usepackage{algorithm}
\usepackage{algorithmic}

\usepackage{booktabs} 
\usepackage{multirow} 
\usepackage{xcolor}   
\usepackage{graphicx} 
\usepackage{caption}  
\usepackage{array}    
\newtheorem{theorem}{Theorem}
\begin{document}
\title{RUQuant: Towards Refining Uniform Quantization for Large Language Models}

\author{Han Liu}
\affiliation{
\institution{Dalian University of Technology}
\city{Dalian}
\country{China}}
\email{liu.han.dut@gmail.com}

\author{Haotian Gao}
\affiliation{
  \institution{Dalian University of Technology}
  \city{Dalian}
  \country{China}}
\email{haotian.dlut@gmail.com}

\author{Changya Li}
\affiliation{
  \institution{Dalian University of Technology}
  \city{Dalian}
  \country{China}}
\email{lichangya.dut@gmail.com}

\author{Feng Zhang}
\affiliation{
  \institution{Peking University}
  \city{Beijing}
  \country{China}}
\email{fengzhangyvonne@gmail.com}

\author{Xiaotong Zhang}
\authornote{Corresponding author.}
\affiliation{
\institution{Dalian University of Technology}
\city{Dalian}
\country{China}}
\email{zxt.dut@hotmail.com}

\author{Wei Wang}
\affiliation{
  \institution{Macao Polytechnic University}
  \city{Macao}
  \country{China}}
\email{weiwang@mpu.edu.mo}

\author{Hong Yu}
\affiliation{
  \institution{Dalian University of Technology}
  \city{Dalian}
  \country{China}}
\email{hongyu@dlut.edu.cn}

\renewcommand{\shortauthors}{Han Liu et al.}

\begin{abstract}
The increasing size and complexity of large language models (LLMs) have raised significant challenges in deployment efficiency, particularly under resource constraints. Post-training quantization (PTQ) has emerged as a practical solution by compressing models without requiring retraining. While existing methods focus on uniform quantization schemes for both weights and activations, they often suffer from substantial accuracy degradation due to the non-uniform nature of activation distributions. In this work, we revisit the activation quantization problem from a theoretical perspective grounded in the Lloyd-Max optimality conditions. We identify the core issue as the non-uniform distribution of activations within the quantization interval, which causes the optimal quantization point under the Lloyd-Max criterion to shift away from the midpoint of the interval. To address this issue, we propose a two-stage orthogonal transformation method, RUQuant. In the first stage, activations are divided into blocks. Each block is mapped to uniformly sampled target vectors using composite orthogonal matrices, which are constructed from Householder reflections and Givens rotations. In the second stage, a global Householder reflection is fine-tuned to further minimize quantization error using Transformer output discrepancies. Empirical results show that our method achieves near-optimal quantization performance without requiring model fine-tuning: RUQuant achieves 99.8\% of full-precision accuracy with W6A6 and 97\% with W4A4 quantization for a 13B LLM, within approximately one minute. A fine-tuned variant yields even higher accuracy, demonstrating the effectiveness and scalability of our approach.
\end{abstract}
\maketitle
\begin{figure*}[!t]
    \label{fig1:all}
    \centering
    \subfigure[Original]{
        \label{fig1:a}
        \includegraphics[width=0.45\columnwidth]{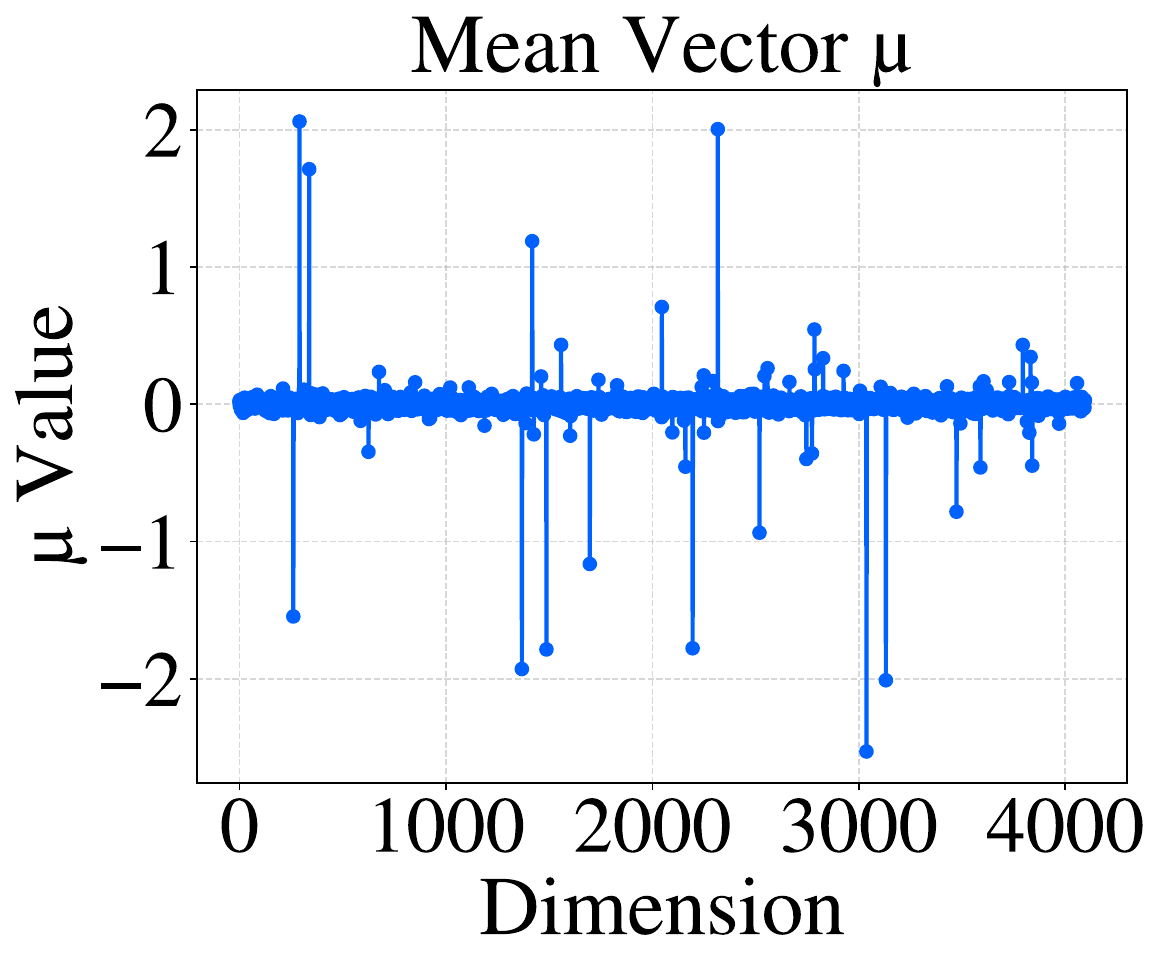}
    }
    \subfigure[RUQuant processed]{
    \label{fig1:b}
    \includegraphics[width=0.45\columnwidth]{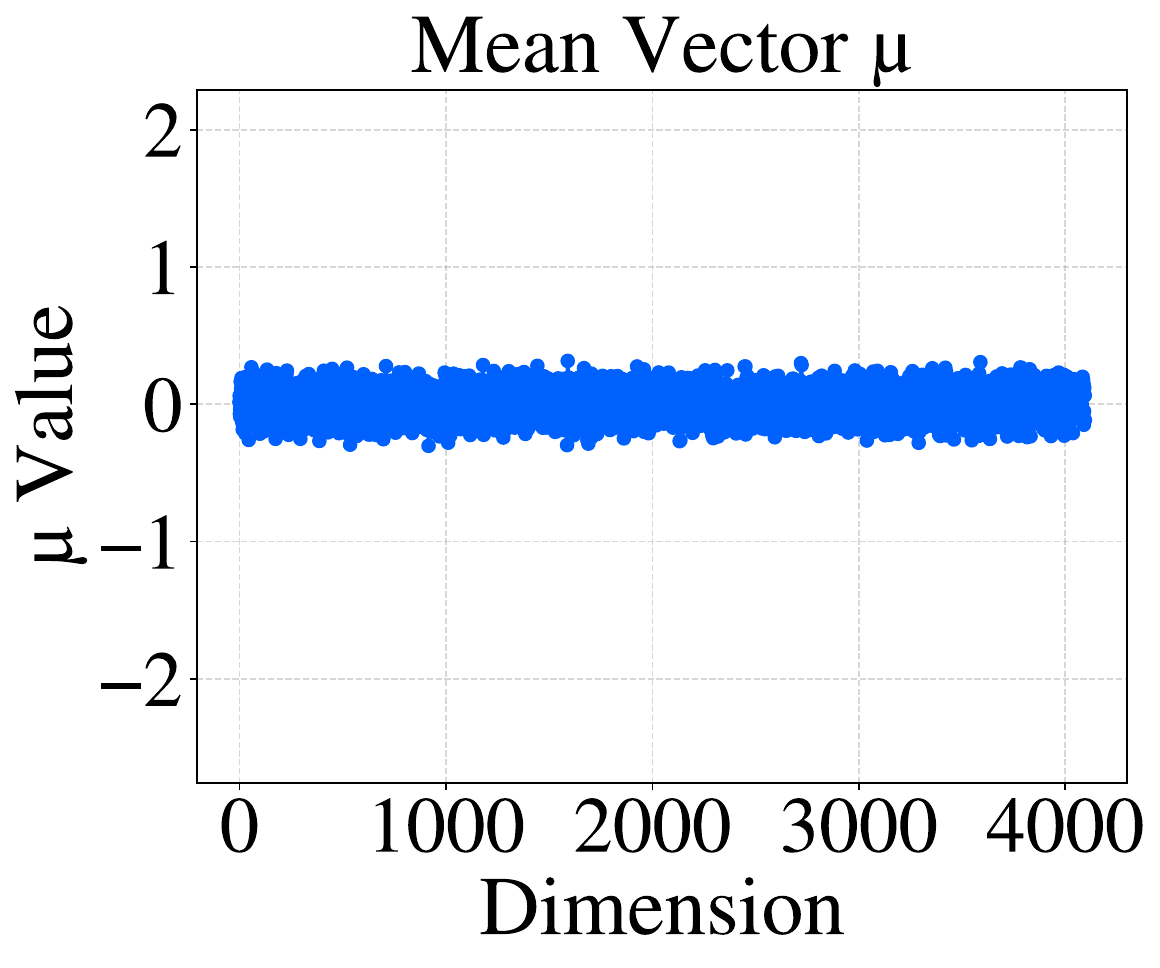}
    }
    \subfigure[Original]{
        \label{fig1:c}
        \includegraphics[width=0.51\columnwidth]{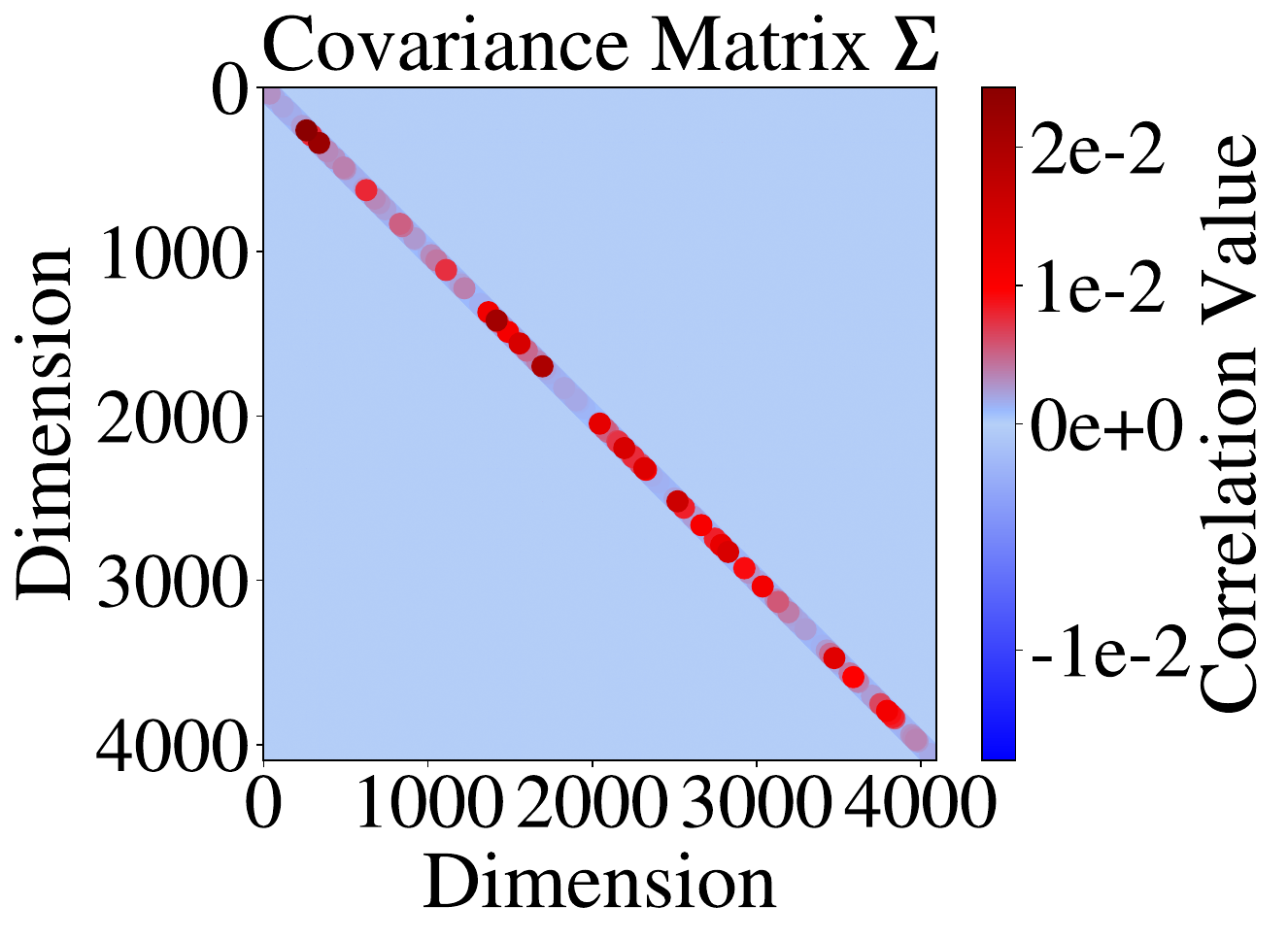}
    }
    \subfigure[RUQuant processed]{
        \label{fig1:d}
        \includegraphics[width=0.51\columnwidth]{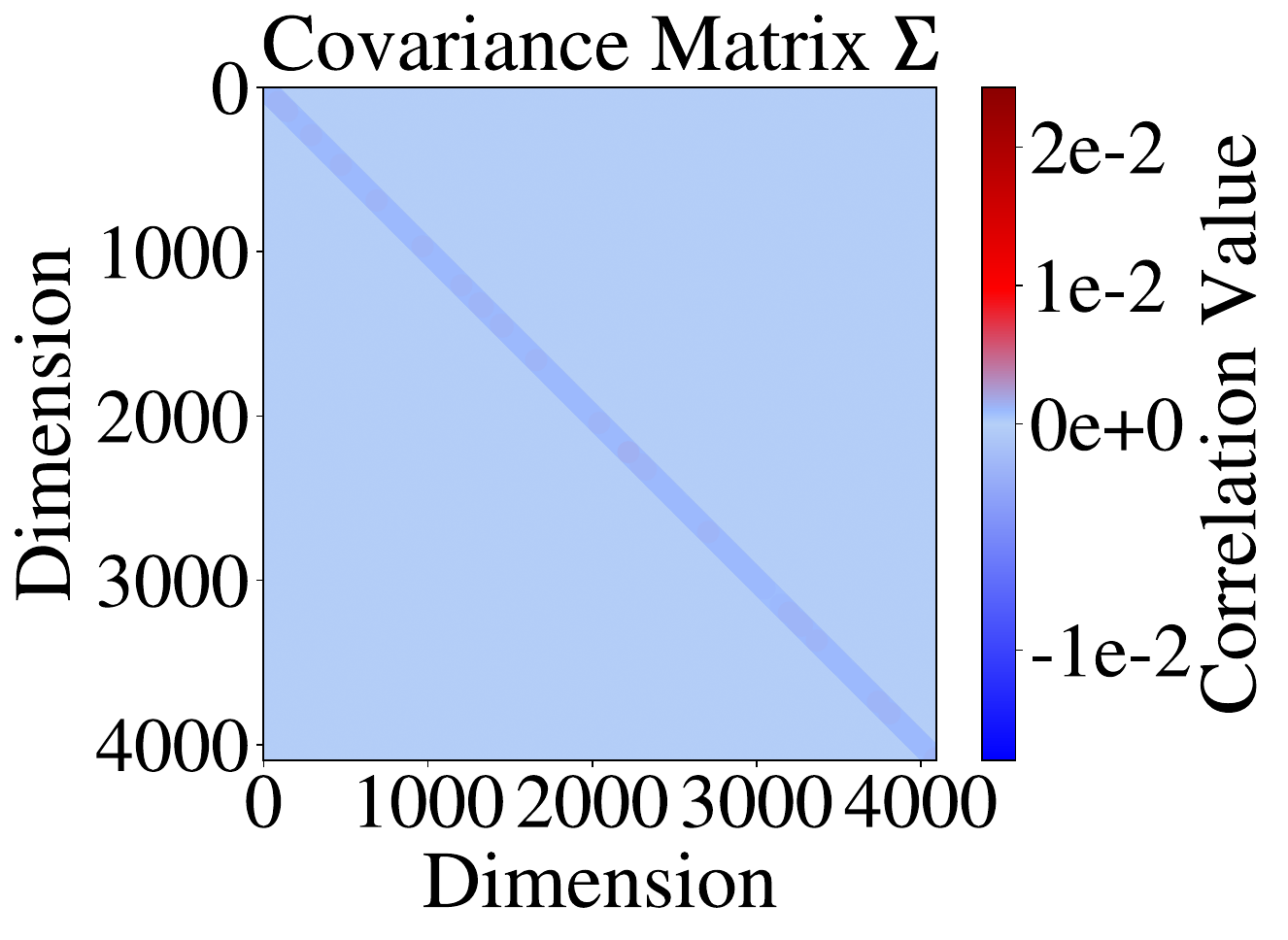}
    }
    \caption{The mean and covariance visualizations of activations before and after RUQuant processing. (a) Original mean vector. (b) Mean vector after RUQuant processing. (c) Original covariance matrix. (d) Covariance matrix after RUQuant processing.}
    \label{fig1:all}
\end{figure*}

\section{Introduction}
Large language models (LLMs) \cite{workshop2022bloom,radford2019language,devlin2018bert,liu2019roberta} based on the Transformer architecture \cite{vaswani2017attention}, achieve remarkable success across a wide range of tasks. With the scaling of these models, the computational and memory demands make deployment increasingly challenging, especially under resource-constrained conditions. These challenges underscore the importance of model compression techniques. Among them, post-training quantization (PTQ) is particularly effective, as it compresses models without requiring costly retraining. 

Recent PTQ works, such as GPTQ \cite{GPTQ}, SpQR \cite{SpQR}, QuIP \cite{Quip}, and AWQ \cite{AWQ}, focus on weight-only quantization for LLMs, converting full-precision weights into low-bit formats to minimize memory usage while achieving nearly lossless 4-bit quantization. To further improve inference efficiency, several works \cite{Smoothquant,Affinequant,Duquant} shift toward weight-activation quantization, which quantizes both weights and activations into low-bit representations for accelerating matrix multiplications. To ensure hardware efficiency and simplify implementation, most of these methods adopt uniform quantizers. However, empirical evidence shows that activations often exhibit highly non-uniform distributions \cite{Duquant}, which can result in significant quantization errors under uniform quantization schemes. While some approaches attempt to address this issue by suppressing outliers and reducing the dynamic range of activations—using techniques such as clipping \cite{OMniquant}, outlier channel scaling \cite{Smoothquant}, or parameter search \cite{flatquant,spinquant}—these methods are primarily heuristic and cannot systematically resolve the fundamental mismatch between non-uniform activation distributions and uniform quantizers.

In this work, we revisit the problem from a theoretical perspective, based on the Lloyd-Max quantization conditions, which characterize the optimality criteria for quantizer design—the decision boundary condition and the centroid condition, i.e., Eq. \eqref{eq2} and Eq. \eqref{eq3}. Our analysis reveals that the performance degradation of uniform quantizers arises primarily from the mismatch between non-uniform activation distributions and the uniform quantization grid. Specifically, in non-uniform distributions, the optimal quantization points (centroids) deviate from the midpoints of quantization intervals, thereby violating the assumptions underlying uniform quantization and resulting in increased quantization error. To overcome this problem, we propose a simple yet effective quantization method, named RUQuant, which facilitates the uniform quantizer by transforming non-uniform activation distributions into approximately uniform ones. This transformation enables the uniform quantizer to achieve performance closer to the theoretical optimum defined by the Lloyd-Max conditions.

Specifically, our proposed  RUQuant is a two-stage method that refines uniform quantization through orthogonal transformations, to improve the efficiency and accuracy when deploying LLMs. In the first stage, to reduce parameter overhead, we partition activations into small blocks, and sample a target vector using a uniform distribution generator for each block. We then construct a composite orthogonal matrix—combining Householder reflections and Givens rotations—by accurately computing a closed-form transformation, such that the original activation vectors are mapped precisely to their corresponding target directions. In the second stage, to achieve uniformity between blocks, we introduce a global refinement step. A single Householder reflection vector is initialized via uniform sampling, and then fine-tuned by optimizing the discrepancy between the outputs of Transformer blocks before and after quantization to minimize quantization error. This step further refines the reflection hyperplane, allowing it to better align with the structure of the activation space.

The main contributions of this work can be summarized as follows: 1) We propose RUQuant, a novel two-stage orthogonal transformation method that effectively transforms non-uniform activation distributions into approximately uniform ones, shown in Figure \ref{fig1:all}, enabling more accurate uniform quantization without requiring model retraining.
2) Theoretically, we revisit activation quantization through the lens of Lloyd-Max optimality conditions, revealing the core cause of quantization errors in uniform schemes.
3) Experimentally, RUQuant demonstrates superior performance on LLMs, specifically on a 13B-scale model, achieving near full-precision accuracy in one minute without fine-tuning\footnote{The source code is available at: https://github.com/RUQuant-code/RUQuant.}.

\begin{figure*}[!t]
    \centering
    \includegraphics[width=0.9\linewidth]{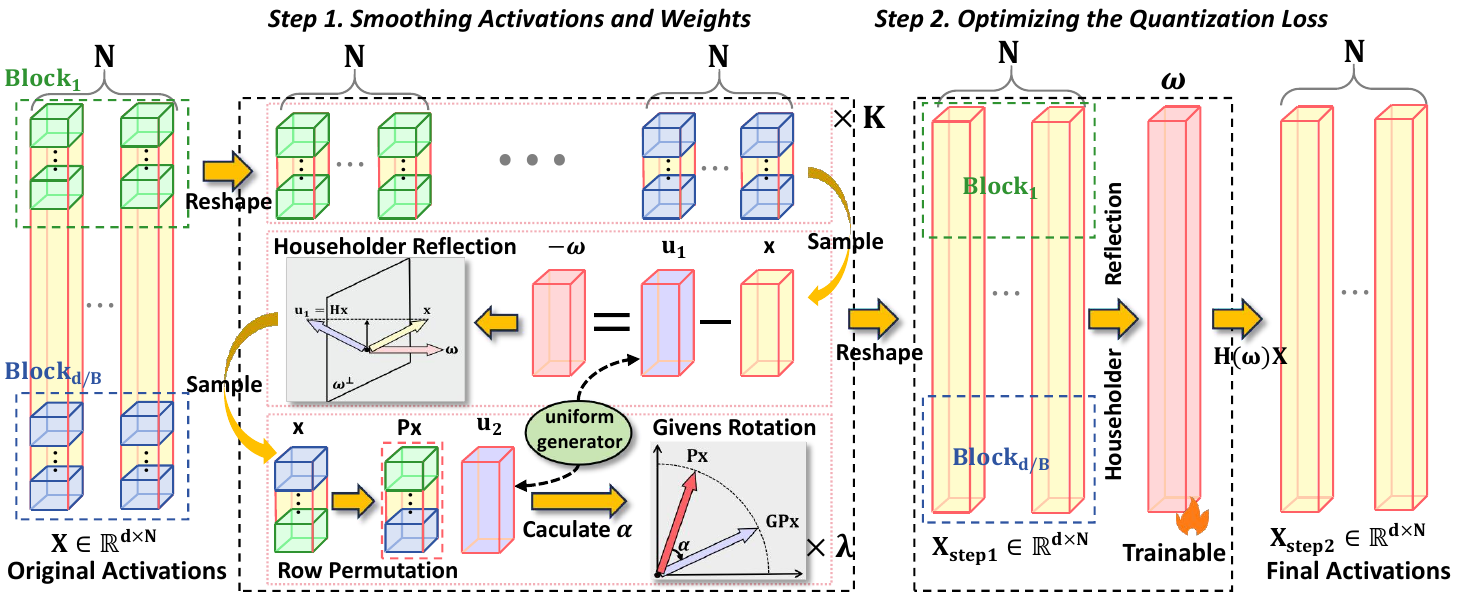}
    \caption{The overall framework of RUQuant (we omitted the zigzag permutation process in the figure). Original activations with size \( d \times N \) are reshaped into \( B \times \frac{dN}{B}\), all column vectors sharing a common rotation matrix. In Step 1, sampled vector $\mathbf{x} \in \mathbb{R}^{B}$ is transformed using Householder and Givens rotations based on uniformly generated vectors \( \mathbf{u}_1 \) and \(\mathbf {u}_2 \), then reconstructed. In Step 2, a trainable Householder matrix is optimized to minimize quantization loss. }
    \label{fig:over}   
\end{figure*}

\section{Related Work}
\label{related work}
Quantization reduces model storage and computation by converting floating-point values into lower-bit representations \cite{zongshu,zongshu2}. Post-training quantization (PTQ) is a practical and widely adopted approach that applies quantization after training, without modifying the training pipeline or requiring model retraining \cite{nagel2020adaround,wu2024ptq4dit,zhao2024mixdq,septq,gptaq,leanquant}. Its simplicity and low computational cost make PTQ especially appealing for large language models (LLMs).

To further decrease memory footprint and computational cost, recent efforts extend PTQ to joint weight–activation quantization. However, activation quantization remains challenging due to highly dynamic input distributions and sensitivity to outliers \cite{llmqat}. To address this, a series of methods introduce equivalent transformation modules to reshape weight and activation distributions while preserving model outputs. SmoothQuant \cite{Smoothquant} first proposes a channel-wise scaling strategy that shifts activation outliers into weights, reducing activation quantization error. Building on this idea, OmniQuant \cite{OMniquant} incorporates learnable scaling and bias terms to better smooth weight distributions. AffineQuant \cite{Affinequant} further generalizes this approach by employing a learnable affine transformation matrix to minimize quantization error in a more expressive manner. DuQuant \cite{Duquant} introduces dual transformations to suppress both large-magnitude activations and activation outliers, achieving improved robustness under activation quantization. QuaRot \cite{quarot} leverages randomized Hadamard rotations to remove activation outliers and enable end-to-end 4-bit quantization, while SpinQuant \cite{spinquant} mitigates randomness-induced instability via learnable Cayley-SGD rotations. Most recently, DFRot \cite{DFRot} identifies tokens with massive activations and optimizes rotations specifically for them, achieving improved accuracy in 4-bit settings. Similar equivalent transformation strategies have also been explored beyond quantization, such as in structured pruning and low-rank decomposition \cite{SliceGPT, SVDLLM}. These extensions further highlight the generality of transformation-based approaches across model compression paradigms.

\section{Motivation}
A $b$-bit uniform quantizer divides the real axis into \(N = 2^b\) equal intervals and uses the midpoints of these intervals as quantization points. For an input vector \(\mathbf{x}\), the quantized low-bit integer is:
\begin{equation}
\mathbf{x}_q = \left( \left\lfloor \frac{\mathbf{x}}{s} \right\rceil + z \right), \quad 
s = \frac{\max(\mathbf{x}) - \min(\mathbf{x})}{2^b - 1}, \quad 
z = -\left\lfloor \frac{\min(\mathbf{x})}{s} \right\rceil,
\end{equation}
where \(s\) is the step size and \(z\) is the zero-point. This fixed-step scheme enables efficient integer operations and thus fast inference. While fixed-step uniform quantization offers advantages in computational efficiency and implementation simplicity, its performance heavily depends on a key assumption that the input data is uniformly distributed within the quantization range. 

However, when quantizing weights and activations simultaneously, this assumption often fails to hold. In LLMs, activations exhibit highly non-uniform distributions across dimensions, which leads to suboptimal quantization outcomes. To illustrate this, we visualize the statistical properties of activations. As shown in Figure~\ref{fig1:a}, the activation means vary significantly across dimensions, ranging approximately from \([-2, 2]\), indicating an uneven distribution of values. Figure~\ref{fig1:c} shows that the covariance matrix is nearly diagonal: most dimensions exhibit low variance, while a few show large fluctuations. This suggests that the activations are weakly correlated but have highly imbalanced scales across channels.

\paragraph{\textbf{Empirical observation.}} For a uniform quantizer, such outlier values in the activation distribution can significantly expand the overall quantization range. Because the fixed step size of a uniform quantizer must cover the outliers, most activation values concentrated near the center are quantized with lower precision, resulting in significant quantization error. This issue stems directly from the non-uniformity of the activation distribution, and this mismatch inherently limits the performance of uniform quantizers. This empirical observation aligns with the theoretical limitations of uniform quantization, which we explore in the next paragraph.

\paragraph{\textbf{Theoretical analysis.}} Given a input vector \(\mathbf{x}\) with a probability density function \(p(x)\), we aim to design a \(b\)-bit quantizer that maps \(\mathbf{x}\) to \(N = 2^b\) quantization points \(\{q_k\}_{k=0}^{N-1}\). The input space is partitioned into \(N\) intervals \(\{[a_k, a_{k+1}]\}_{k=0}^{N-1}\). The goal of the quantization is to minimize the mean squared error (MSE):
\begin{equation}
\label{eq1}
    E = \sum_{k=0}^{N-1} \int_{a_k}^{a_{k+1}} (x - q_k)^2 p(x) \, dx.
\end{equation}
To minimize Eq. \eqref{eq1}, we differentiate it with respect to $a_k$ and $q_k$, and solve for the optimal values. The resulting expressions satisfy the Lloyd-Max conditions \cite{LloyedMax} (Please refer to Appendix~\ref{appA} for a detailed derivation):

\textbf{(1) Decision boundary condition}: The interval boundaries \(a_k\) are located at the midpoints between adjacent quantization points,
\begin{equation}
\label{eq2}
a_k = \frac{q_{k-1} + q_k}{2}.
\end{equation}

\textbf{(2) Centroid condition}: The quantization points \(q_k\) are the conditional expectations of the corresponding intervals,
\begin{equation}
\label{eq3}
q_k = \frac{\int_{a_k}^{a_{k+1}} x p(x) \, dx}{\int_{a_k}^{a_{k+1}} p(x) \, dx}.
\end{equation}

Obviously, the uniform quantizer naturally satisfies the decision boundary condition Eq. \eqref{eq2}, because each decision boundary is the midpoint of adjacent quantization points. However, due to the non-uniformity of \(p(x)\) within the quantization intervals, the optimal centroid of each interval (defined by Eq. \eqref{eq3}) may deviate from the midpoint, thus making uniform quantization no longer optimal.

\begin{algorithm}[!t]
\caption{Generate Householder orthogonal matrix}
\label{algorithm1}
\parbox[t]{\linewidth}{\textbf{Input}: Activation vector $\mathbf{x} \in \mathbb{R}^d$}\\
\parbox[t]{\linewidth}{\textbf{Output}: Householder orthogonal matrix $\mathbf{H}\in \mathbb{R}^{d\times d}$}\\
\begin{algorithmic}[1]
    \STATE $\gamma \gets \|\mathbf{x}\|_2$
    \hfill // Calculate input vector norm
    \STATE $\mathbf{u} \gets \mathcal{U}(-1,1)^d$
    \hfill // Get uniform distribution in $[-1,1]^d$
    \STATE $\mathbf{u} \gets \gamma \cdot \mathbf{u}/\|\mathbf{u}\|_2$
    \hfill // Match input vector norm
    \STATE  $\boldsymbol{\omega} \gets \mathbf{x}-\mathbf{u}$
    \hfill // Construct reflection axis
    \STATE $\mathbf{H} \gets \mathbf{I}_d - \frac{2}{\boldsymbol{\omega}^\top \boldsymbol{\omega}}\boldsymbol{\omega} \boldsymbol{\omega} ^\top$
    \hfill // Get the Householder matrix by Eq. \eqref{eq9}
\end{algorithmic}
\end{algorithm}

Based on both empirical observation and theoretical analysis, our goal is to find a transformation $F$ that maps $\mathbf{x}$ to a distribution that is closer to uniform. This transformation aims to align the quantization levels of the uniform quantizer with the optimal quantization points, thereby enabling the quantizer to approach the theoretical optimum characterized by Lloyd-Max quantization theory.

\section{The Proposed Method}
The RUQuant method comprises two key steps: uniformization of activation and weight distributions, and optimization of quantization error. In the first step, we utilize two types of orthogonal transformations—Householder reflections and Givens rotations—to map activations to a uniform distribution, thereby minimizing the loss introduced by uniform quantizers. In the second step, we design Learnable Householder matrices and fine-tune the orthogonal transformations by minimizing the discrepancy between model outputs before and after quantization, which improves the quantization accuracy. The overall framework of the RUQuant algorithm is illustrated in Figure \ref{fig:over}.

\subsection{Smoothing Activations and Weights}
\label{sec3.2}
To uniformize activations and weights simultaneously, we consider the transformation: $\mathbf{Y} = \mathbf{W} \mathbf{A}^{-1} \mathbf{A} \mathbf{X}$, where computing the inverse can be costly. Therefore, we let \(\mathbf{A}\) be an orthogonal matrix so that \(\mathbf{A}^{-1} = \mathbf{A}^\top\), which reduces computational overhead and effectively altering the distribution. This form of equivalent transformation has been demonstrated effective in various prior works \cite{Affinequant,Duquant}. In this section, we introduce a closed-form orthogonal matrix that maps activations into more uniformly distributed directions. Accordingly, the linear layer in LLMs can be equivalently rewritten as: $\mathbf{Y} = \mathbf{W} \mathbf{X} = \mathbf{W} \mathbf{Q_c}^\top \mathbf{Q_c} \mathbf{X}$, where \(\mathbf{W} \in \mathbb{R}^{m \times d}\), \(\mathbf{X} \in \mathbb{R}^{d \times N}\), \(\mathbf{Q_c} \in \mathbb{R}^{d \times d}\), with \(d\) denoting the feature dimension and \(N\) the number of samples.

The uniformization step can be understood as follows: we uniformly sample target vectors and map an activation vector \(\mathbf{x} \in \mathbb{R}^{d}\) to these target directions \(\mathbf{u} \in \mathbb{R}^d\) in order to uniformize the distribution of activations. This mapping is realized through a composite orthogonal transformation composed of multiple Householder reflections and Givens rotations. To improve the expressive capacity and stability of the transformation, we perform multiple rounds of uniform sampling rather than a single one.

Specifically, a Householder reflection transformation \(\mathbf{H} \in \mathbb{R}^{d \times d}\) can be uniquely determined by a vector \(\boldsymbol{\omega} \in \mathbb{R}^d\):
\begin{equation}
\label{eq9}
\mathbf{H}(\boldsymbol{\omega}) = \mathbf{I}_d - \frac{2}{\boldsymbol{\omega}^\top \boldsymbol{\omega}} \boldsymbol{\omega} \boldsymbol{\omega}^\top.
\end{equation}
For any two vectors \(\mathbf{x}, \mathbf{y} \in \mathbb{R}^d\) with \(\|\mathbf{x}\|_2 = \|\mathbf{y}\|_2\) and \(\mathbf{x} \neq \mathbf{y}\), there must exist a Householder matrix such that \(\mathbf{y} = \mathbf{H}(\boldsymbol{\omega}) \mathbf{x}\), where \(\boldsymbol{\omega} = \mathbf{x} - \mathbf{y}\). Therefore, we can map each activation vector to a uniform vector \(\mathbf{u}\) using a Householder matrix. The specific procedure is described in Algorithm \ref{algorithm1}.

In the 2D plane \(\mathbb{R}^2\), a Givens rotation matrix can be uniquely determined by a single angle \(\alpha\):
\begin{equation}
\mathbf{\Gamma}(\alpha) = \begin{pmatrix}
\cos\alpha & \sin\alpha \\
-\sin\alpha & \cos\alpha
\end{pmatrix}.
\end{equation}
Inspired by the RoPE (Rotary Position Embedding) method \cite{RoPE}, we extend this two-dimensional rotation to higher dimensions by grouping the elements of a high-dimensional vector into pairs and rotating each pair separately. Thus, the rotation matrix \(\mathbf{G} \in \mathbb{R}^{d \times d}\) can be uniquely determined by a vector \(\mathbf{a} \in \mathbb{R}^{\frac{d}{2}}\):
\begin{equation}
\label{eq11}
\mathbf{G}(\mathbf{a}) = \text{BlockDiag}(\mathbf{\Gamma}(\mathbf{a}_1), \mathbf{\Gamma}(\mathbf{a}_2),\dots, \mathbf{\Gamma}(\mathbf{a}_{\frac{d}{2}})).
\end{equation}
For any two vectors \(\mathbf{p}, \mathbf{v} \in \mathbb{R}^2\) with \(\|\mathbf{p}\|_2 = \|\mathbf{v}\|_2\) and \(\mathbf{p} \neq \mathbf{v}\), there must exist a Givens matrix such that \(\mathbf{v} = \mathbf{\Gamma}(\mathbf{a}_i)\mathbf{p}\), and the rotation angle \(\mathbf{a}_i\) can be computed as:
\begin{equation}
\label{eq12}
\cos \mathbf{a}_i = \frac{\mathbf{v}_1 \mathbf{p}_1 + \mathbf{v}_2 \mathbf{p}_2}{\mathbf{p}_1^2 + \mathbf{p}_2^2}, \quad
\sin \mathbf{a}_i = \frac{\mathbf{v}_1 \mathbf{p}_2 - \mathbf{v}_2 \mathbf{p}_1}{\mathbf{p}_1^2 + \mathbf{p}_2^2}.
\end{equation}

Since \(\mathbf{G}\) is a block diagonal matrix with block size 2, it only smooths pairs of consecutive values, which are often already uniformly distributed. To address this, we introduce a random permutation matrix \(\mathbf{P} \in \mathbb{R}^{d \times d}\) with \(\mathbf{P}^\top \mathbf{P} = \mathbf{I}\) to shuffle the activation vector \(\mathbf{x}\) before rotation. By alternating random permutations and block rotations, we form an orthogonal matrix \(\mathbf{R}\). This process is repeated \(\lambda\) times. The specific procedure is described in Algorithm \ref{algorithm2}.
\begin{algorithm}[!t]   
    \caption{Generate orthogonal matrix by Givens and permutations}
    \label{algorithm2}
    \parbox[t]{\linewidth}{\textbf{Input}: Activation vector $\mathbf{x} \in \mathbb{R}^d$, Number of permutations $\lambda$}\\
    \parbox[t]{\linewidth}{\textbf{Output}:  Orthogonal matrix $\mathbf{R}\in \mathbb{R}^{d\times d}$}\\
    \begin{algorithmic}[1]
        \STATE $\mathbf{R} \gets\mathbf{I}_d$ 
        \hfill // Initialize the orthogonal matrix 
        \STATE $\mathbf{a} \gets\mathbb{R}^{\frac{d}{2}}$ 
        \hfill // Initialize angle vector
        \FOR{$i = 1$ to $\lambda$}
            \STATE $\mathbf{P}\gets\text{Random permutation matrix}$
            \STATE $\mathbf{x}\gets\mathbf{P}\mathbf{x}$ 
            \hfill 
            \STATE $\mathbf{R}\gets\mathbf{P}\mathbf{R}$
            \hfill // Update the orthogonal matrix 
            \STATE $\mathbf{u} \gets \mathcal{U}(-1,1)^d$ 
            \hfill // Get uniform distribution in $[-1,1]^d$
            \FOR{$j = 1$ to $n/2 $}
                \STATE $\mathbf{p} \gets \mathbf{x}[2j:2j+2]$
                \hfill // Extract a sub-vector
                \STATE $\mathbf{v} \gets \mathbf{u}[2j:2j+2]$
                \STATE $\gamma \gets \|\mathbf{p}\|_2$
                \hfill // Calculate the norm of the sub-vector 
                \STATE $\mathbf{v} \gets \gamma \cdot \mathbf{v}/\|\mathbf{v}\|_2$
                \hfill // Match sub-vector norm
                \STATE \(\mathbf{a}_j \gets \)Calculate $\mathbf{a}_j$ by Eq. (\ref{eq12})
            \ENDFOR
            \STATE \( \mathbf{G}(\mathbf{a})\gets\)  Calculate $\mathbf{G}$ by Eq. (\ref{eq11})
            \STATE $\mathbf{R} \gets \mathbf{G}\mathbf{R}$
            \hfill // Update the orthogonal matrix 
        \ENDFOR
    \end{algorithmic}
\end{algorithm}

For each sampled activation vector \(\mathbf{x}\), we compute a reflection orthogonal matrix and a rotation orthogonal matrix according to Algorithm \ref{algorithm1} and Algorithm \ref{algorithm2}, respectively. To enhance stability, we adopt an iterative sampling approach to generate \(\mathbf{Q}_c\). Let \(K\) denote the number of sampling iterations. We perform \(K\) iterations of these two algorithms, resulting in
\begin{equation}
\label{eq14}
    \mathbf{Q}_c^\top = (\mathbf{R}_1 \mathbf{H}_1)^\top (\mathbf{R}_2 \mathbf{H}_2 )^\top\cdots (\mathbf{R}_K \mathbf{H}_K)^\top.
\end{equation}

\subsubsection{Smoothing Effect on Activation}
Based on the method above, we can map a sampled activation vector to a uniformly distributed space. However, it is impractical to compute a dedicated projection matrix for every activation vector individually. Instead, we investigate whether the orthogonal projection matrix derived from a single sampled vector \(\mathbf{x}\) from the activation matrix \(\mathbf{X}\) can generalize to smooth the entire activation matrix. As shown in Theorem \ref{theorem1}, the constructed orthogonal matrix is capable of effectively smoothing \(\mathbf{X}\). The complete derivation is provided in Appendix \ref{appB}.

\begin{theorem}[Smoothing effect of orthogonal transformation on activations]
\label{theorem1}
Let \(\mathbf{x} \in \mathbb{R}^d\) be an activation vector sampled from \(\mathcal{N}(\boldsymbol{\mu}, \boldsymbol{\Sigma})\), and let \(\mathbf{Q} \in \mathbb{R}^{d \times d}\) be an orthogonal matrix such that \(\mathbf{Qx} = \mathbf{u}\), where \(\mathbf{u}\) is a fixed vector sampled from a uniform distribution. Then, for any new activation \(\mathbf{x}_{\text{new}} \sim \mathcal{N}(\boldsymbol{\mu}, \boldsymbol{\Sigma})\), the transformed vector satisfies \(\mathbf{Q}\mathbf{x}_{\text{new}} \sim \mathcal{N}(\mathbf{Q}\boldsymbol{\mu}, \mathbf{Q}\boldsymbol{\Sigma}\mathbf{Q}^\top)\). Moreover, the transformation approximately satisfies
\begin{equation}
\label{eqx}
\mathbf{Q}\boldsymbol{\mu} \approx \mathbf{u}, \quad \mathbb{E}[\mathbf{Q}\boldsymbol{\Sigma}\mathbf{Q}^\top] = \frac{\operatorname{tr}(\boldsymbol{\Sigma})}{d} \mathbf{I},
\end{equation}
where \(\mathbf{I}\) is the identity matrix.
\end{theorem}

We observe that the value of \(\text{tr}(\boldsymbol{\Sigma})/d\) is very small, indicating that the distribution of \(\mathbf{x}\) after the transformation by the orthogonal matrix \(\mathbf{Q}_c\) is centered around the uniform vector \(\mathbf{u}\), with small isotropic fluctuations in each dimension. We show the mean vector and covariance matrix of \(\mathbf{Q}_c\mathbf{X}\) in Figure \ref{fig1:all}. It can be seen that the mean of \(\mathbf{Q}_c\mathbf{X}\) (Figure \ref{fig1:b}) fluctuation range is very small, showing a relatively uniform trend, and the covariance matrix (Figure \ref{fig1:d}) satisfies the isotropy as shown in Eq. (\ref{eqx}). It can be seen that the transformed activations are concentrated around the mean vector and the fluctuation range is very small.

\subsubsection{Smoothing Effect on Weight}
The previous strategy focuses on smoothing the activation matrix \(\mathbf{X}\) by applying an orthogonal transformation \(\mathbf{Q}\). However, due to the equivalence of the transformation \(\mathbf{W}\mathbf{X} = (\mathbf{W}\mathbf{Q}^\top)(\mathbf{Q}\mathbf{X})\), the same orthogonal matrix \(\mathbf{Q}\) must be applied to the weight matrix \(\mathbf{W}\) from the right. While \(\mathbf{Q}\) is constructed solely based on the distribution of \(\mathbf{X}\), we empirically observe that this transformation also has a smoothing effect on \(\mathbf{W}\). Theoretically, the following theorem shows that applying the transpose of the activation-based orthogonal matrix \(\mathbf{Q}\) to the weight matrix effectively reduces its distributional anisotropy. The full derivation is provided in Appendix~\ref{appC}.

\begin{theorem}[Smoothing effect of orthogonal transformation on weight]
\label{theorem2}
Let \(\mathbf{Q} \in \mathbb{R}^{d \times d}\) be an orthogonal matrix constructed from activation statistics (Section~\ref{sec3.2}), and let a weight matrix \(\mathbf{W} \in \mathbb{R}^{m \times d}\) be transformed as \(\mathbf{W} \leftarrow \mathbf{W} \mathbf{Q}^\top\). Assume each row vector \(\mathbf{w}^\top\) of \(\mathbf{W}\) is independently drawn from \(\mathcal{N}(\boldsymbol{\mu}_W, \boldsymbol{\Sigma}_W)\), where \(\boldsymbol{\mu}_W \in \mathbb{R}^d\) and \(\boldsymbol{\Sigma}_W \in \mathbb{R}^{d \times d}\). Then the transformed vector \(\mathbf{Q} \mathbf{w}\) satisfies $\mathbf{Q} \mathbf{w} \sim \mathcal{N}(\mathbf{Q} \boldsymbol{\mu}_W, \mathbf{Q} \boldsymbol{\Sigma}_W \mathbf{Q}^\top)$. Moreover, the transformation
approximately satisfies
\begin{equation}
\label{eqw}
\mathbf{Q} \boldsymbol{\mu}_W \approx \mathbf{0},  \quad
\mathbb{E}[\mathbf{Q} \boldsymbol{\Sigma}_W \mathbf{Q}^\top] = \frac{\operatorname{tr}(\boldsymbol{\Sigma}_W)}{d} \mathbf{I},
\end{equation}
where \(\mathbf{I}\) is the identity matrix.
\end{theorem}

In fact, our experimental observations also corroborate Eq \eqref{eqw}: our $\mathbf{Q}_c$ not only effectively smooths the activations, but also smooths the weights. Although the original weights $\mathbf{W}$ are already relatively uniform, after the transformation by $\mathbf{WQ}_c^\top$, the weight vectors become more evenly distributed and exhibit smaller variations across dimensions. Additional visualizations are provided in Appendix~\ref{appD}.

\subsubsection{Block-Wise Rotation}
\label{sec3.2.2}
Although our method is very fast to directly construct the entire orthogonal matrix, there are two problems: 1) The time complexity of multiplying the orthogonal matrix \(\mathbf{Q}_c \in \mathbb{R}^{d\times d}\) and the activation matrix \(\mathbf{X} \in \mathbb{R}^{d \times N}\) is \(O(d^2N)\), which will greatly increase the inference delay of the model. 2) The number of parameters of the orthogonal matrix \( \mathbf{Q}_c\) is \(d^2\), and storing \( \mathbf{Q}_c\) will generate a huge memory overhead. Like DuQuant \cite{Duquant}, we adopted the block-wise rotation and zigzag permutation strategy.

(1) \textbf{Block-Wise Processing}: We approximate the rotation matrix \(\mathbf{Q}_c\) in a block manner. When the block size is $B$, we have:
\begin{equation}
\label{eq15}
\mathbf{\hat{Q}_c}=\text{BlockDiag}(\mathbf{Q}_{b_1},\cdots, \mathbf{Q}_{b_{d/B}}),
\end{equation}
where \( \mathbf{Q}_{b_i} \in \mathbb{R}^{B \times B} \) is the shared square matrix for all blocks, i.e., \( \mathbf{Q}_{b_i} = \mathbf{Q}_b \) for all \( i \)
, and $\mathbf{Q}_b$ is constructed by Eq. (\ref{eq14}). Therefore, the total number of parameters is reduced to \(B^2\), and the computational complexity of multiplying the block-wise orthogonal matrix with the activation matrix is reduced to \(O(BdN)\). In our experiments, we fix \( B = 128 \); for a weight matrix of size \( 4096 \times 4096 \), this accounts for less than 0.1\% of the original parameter count.

(2) \textbf{Zigzag Permutation}: Since the blocks are independent, zigzag permutation can reduce the variance between blocks\cite{Duquant}. Specifically, after the block-wise rotation, we introduce a permutation matrix $\mathbf{Z}\in \mathbb{R}^{d\times d}$ that satisfies $\mathbf{ZZ}^\top=\mathbf{I}$. By shuffling the order of activation elements, the independence between blocks is disrupted, leading to a more uniform distribution across blocks and thereby improving the overall quantization performance.

Finally, the $\mathbf{Q}_c$ of the first step is: $\mathbf{Q}_c=\left(\mathbf{\hat{Q}}_{c}(1) \mathbf{Z}\mathbf{\hat{Q}}_{c}(2)\right)^\top$, where $\mathbf{Z}$ represents the permutation matrix generated by the zigzag permutation, and $\mathbf{\hat{Q}}_{c}(1) ,\mathbf{\hat{Q}}_{c}(2)$ represents the diagonal matrices of the first and second block-wise rotations.
\subsection{Optimizing the Quantization Loss}

\begin{table*}[!t]
\caption{The perplexity results ($\downarrow$) of the LLaMA1, LLaMA2 and LLaMA3 model on the WikiText2 dataset.}
\begin{center}
\tabcolsep=0.45cm
\begin{tabular}{c|l|cccccccc}
\toprule
\textbf{\#Bit}        & \multicolumn{1}{c|}{\textbf{Method}} & \textbf{1-7B} & \textbf{1-13B} & \textbf{1-30B}  & \textbf{2-7B} & \textbf{2-13B} & \textbf{3-8B} \\ 
\midrule
FP16 &-                         & 5.68         & 5.09          & 4.10               & 5.47         & 4.88          & 6.14          \\
\midrule
\multirow{6}{*}{W4A4} 
& SmoothQuant                          & 25.25         & 40.05          & 192.40               & 83.12         & 35.88          & 210.19          \\
                      & OmniQuant                            & 11.26         & 10.87          & 10.33                 & 14.26         & 12.30          & 3.64e3            \\
                      & QLLM                                 & 9.65          & 8.41           & 8.37                   & 11.75         & 9.09           & -          \\           
                    & DuQuant                                 & 6.40          & 5.65           & 4.72                    & 6.28          & 5.42           & 8.56          \\
                      
                      & \cellcolor{brown!10}{RUQuant}           & \cellcolor{brown!10}{6.29}          & \cellcolor{brown!10}{5.55}           & \cellcolor{brown!10}{4.61}                    & \cellcolor{brown!10}{6.17}          & \cellcolor{brown!10}{5.35}           & \cellcolor{brown!10}{ 8.10}           \\
                                            & \cellcolor{brown!10}\textbf{RUQuant{\tiny{+fine-tune}}}           & \cellcolor{brown!10}\textbf{6.14}     
                      & \cellcolor{brown!10}\textbf{5.53}           & \cellcolor{brown!10}\textbf{4.55}
                      & \cellcolor{brown!10}\textbf{6.04}          
                      & \cellcolor{brown!10}\textbf{5.29}           & \cellcolor{brown!10}\textbf{ 7.99}           \\

\bottomrule
\multirow{5}{*}{W6A6} & SmoothQuant                          & 25.25         & 40.05          & 192.40                & 83.12         & 35.88          & 7.07          \\
                      & OmniQuant                            & 11.26         & 10.87          & 10.33                   & 14.26         & 12.30          & 7.24            \\
                      & QLLM                                 & 9.65          & 8.41           & 8.37                   & 11.75         & 9.09           & -           \\
                      & DuQuant                                 & 5.73          & 5.13           & 4.14                     & 5.53          & 4.92           & 6.27           \\
                     
                      & \cellcolor{brown!10}\textbf{RUQuant}          & \cellcolor{brown!10}\textbf{5.71}         & \cellcolor{brown!10}\textbf{5.12}           & \cellcolor{brown!10}\textbf{4.13}                    & \cellcolor{brown!10}\textbf{5.51}         & \cellcolor{brown!10}\textbf{4.91}           & \cellcolor{brown!10}\textbf{6.25}          \\
                     
\bottomrule
\end{tabular}
\end{center}
\label{tab:ppl-wikitext2}
\end{table*}
In this section, to align the quantization loss of the uniform quantizer with the model loss and further improve quantization accuracy, we introduce a Learnable Householder matrix \(\mathbf{Q}_l\). According to Eq.~\eqref{eq9}, a \(d \times d\) reflection matrix \(\mathbf{H}\) can be generated from a \(d\)-dimensional vector \(\boldsymbol{\theta}\). Moreover, the matrix-vector multiplication can be computed efficiently as
$\mathbf{H}\mathbf{X} = \mathbf{X} - \eta \boldsymbol{\theta} (\boldsymbol{\theta}^\top \mathbf{X})$, where \(\eta = 2 / (\boldsymbol{\theta}^\top \boldsymbol{\theta})\), resulting in a computational complexity of \(O(dN)\). Additionally, the parameter count of \(\boldsymbol{\theta}\) is only \(d\) for each activation.
Therefore, introducing a Learnable Householder matrix incurs minimal parameter overhead and low computational cost. Specifically, after Step 1, we obtain \(\mathbf{X}_{step1} = \mathbf{Q}_c \mathbf{X}\). Instead of dividing \(\mathbf{X}_{step1}\) into blocks, we perform a single sampling to initialize \(\boldsymbol{\theta}\) using Algorithm~\ref{algorithm1}, and construct \(\mathbf{Q}_l(\boldsymbol{\theta})\) via Eq.~\eqref{eq9}. We then fine-tune \(\boldsymbol{\theta}\) by minimizing a block-wise loss designed to optimize the quantization accuracy:
\begin{equation}
\label{eq8}
\arg\min_{\boldsymbol{\theta}} \left\| f(\mathbf{W}, \mathbf{X}) - f\left(\mathcal{Q}\left(\mathbf{W} \mathbf{Q}_c^\top \mathbf{Q}_l(\boldsymbol{\theta})^\top \right), \mathcal{Q}\left(\mathbf{Q}_l(\boldsymbol{\theta}) \mathbf{Q}_c \mathbf{X}\right) \right) \right\|_{\text{F}}^2,
\end{equation}
where \(f(\cdot)\) denotes a Transformer block, \(\mathcal{Q}(\cdot)\) is the quantization function, and \(\|\cdot\|_{\text{F}}^2\) denotes the squared Frobenius norm.

\subsection{The Overall Method}
We provide a detailed algorithm in Appendix \ref{APPG}. For each Transformer block \(\mathcal{T}\), the procedure is performed in two main steps. In the first step, we apply Householder and Givens transformations to all linear layers within the block. The input \(\mathbf{X}\) is first partitioned into blocks of size \(B\). Then, all blocks share a common rotation matrix \(\mathbf{Q}_c\), which, through the composite linear transformations of Householder reflections and Givens rotations, precisely maps the activations from a non-uniform to a uniform distribution. To further reduce inter-block variance, we apply a zigzag permutation followed by an additional rotation. In the second step, we initialize the parameters \(\boldsymbol{\theta}\) of a Learnable Householder matrix using Algorithm 1, and fine-tune \(\boldsymbol{\theta}\) by minimizing the difference in outputs of the Transformer block before and after quantization. This fine-tuning process helps mitigate the artifacts introduced by block-wise processing and better aligns the quantization loss with the overall model loss. 

\begin{table*}[!t]
\caption{Zero-shot QA ($\uparrow$) results of LLaMA2 and LLaMA3 models under 4-bit weight-activation quantization.
}
\begin{center}
\tabcolsep=0.3cm
\renewcommand{\arraystretch}{1}
\begin{tabular}{c|l|ccccccc}
\toprule
\textbf{Model}     & \textbf{Method}  & \textbf{PIQA}  & \textbf{ARC-E} & \textbf{ARC-C} & \textbf{BoolQ} & \textbf{HellaSwag} & \textbf{WinoGrande} & \textbf{Avg.}   \\ 
\midrule
\multirow{6}{*}{\begin{tabular}[c]{@{}c@{}}LLaMA2-7B\\ W4A4\end{tabular}} 
& FP16             & 76.88          & 53.54          & 40.53          & 71.13          & 72.96              & 67.25               & 63.72          \\ \cmidrule{2-9} 
& SmoothQuant      & 60.17          & 35.23          & 27.13          & 57.92          & 37.08              & 49.57               & 44.52          \\
& OmniQuant        & 65.61          & 44.28          & 30.38          & 62.66          & 53.51              & 51.85               & 51.38          \\
& QLLM             & 67.68          & 44.40          & 30.89          & -              & 58.45              & 56.59               & 51.60          \\
& DuQuant             & 75.24          & 51.89          & 36.77          & 67.86          & 69.54              & 62.12               & 60.57          \\

& \cellcolor{brown!10}\textbf{RUQuant}           & \cellcolor{brown!10}\textbf{75.84}   & \cellcolor{brown!10}\textbf{52.82}           &\cellcolor{brown!10}\textbf{39.51} 
& \cellcolor{brown!10}\textbf{69.57}  & \cellcolor{brown!10}\textbf{70.06}   &\cellcolor{brown!10}\textbf{63.61}   &\cellcolor{brown!10}\textbf{61.90} \\

& \cellcolor{brown!10}\textbf{RUQuant{\tiny{+fine-tune}}}     & \cellcolor{brown!10}{74.65}          & \cellcolor{brown!10}{51.18}   & \cellcolor{brown!10}{38.65} 
& \cellcolor{brown!10}68.59 & \cellcolor{brown!10}{69.65}    & \cellcolor{brown!10}61.72   & \cellcolor{brown!10}60.74\\
\midrule
\multirow{6}{*}{\begin{tabular}[c]{@{}c@{}}LLaMA2-13B\\ W4A4\end{tabular}} 
& FP16             & 79.05          & 57.91          & 44.20          & 69.02          & 76.60              & 69.69               & 66.08          \\ \cmidrule{2-9} 
& SmoothQuant      & 62.30          & 40.28          & 30.72          & 60.49          & 42.24              & 49.96               & 47.67          \\
& OmniQuant        & 69.80          & 47.22          & 33.79          & 65.47          & 59.34              & 55.49               & 55.19          \\
& QLLM             & 70.46          & 48.48          & 34.39          & -              & 62.80              & 55.41               & 54.31          \\
& DuQuant             & 77.31          & 55.60         & 41.55          & 66.61          & 73.68             & 66.06               & 63.47          \\

& \cellcolor{brown!10}\textbf{RUQuant}     & \cellcolor{brown!10}{78.24}   & \cellcolor{brown!10}55.47  & \cellcolor{brown!10}41.30          & \cellcolor{brown!10}68.20    & \cellcolor{brown!10}\textbf{74.07}        & \cellcolor{brown!10}\textbf{67.64}      & \cellcolor{brown!10}\textbf{64.15}    \\

& \cellcolor{brown!10}\textbf{RUQuant{\tiny{+fine-tune}}}     & \cellcolor{brown!10}\textbf{78.51}          & \cellcolor{brown!10}\textbf{56.14}   & \cellcolor{brown!10}\textbf{41.98} 
& \cellcolor{brown!10}\textbf{68.41} & \cellcolor{brown!10}73.78    & \cellcolor{brown!10}65.98   & \cellcolor{brown!10}64.13\\

\midrule
\multirow{6}{*}{\begin{tabular}[c]{@{}c@{}}LLaMA3-8B\\ W4A4\end{tabular}} 
& FP16             & 80.85          & 77.78          & 53.41          & 81.28          & 79.16              & 72.84              & 74.22         \\ \cmidrule{2-9} 
& SmoothQuant      &54.57          & 31.9          & 24.23          & 52.72          & 31.26              & 51.14               & 40.97         \\
& OmniQuant        & 50.22          & 26.94          & 24.57          & 37.98          & 26.55              &50.20               & 36.08          \\
& DuQuant             & 75.68          & 68.48          & 41.81          & 71.99          & 73.07              & 66.22               & 66.21         \\

& \cellcolor{brown!10}\textbf{RUQuant}     & \cellcolor{brown!10}\textbf{76.93}   & \cellcolor{brown!10}\textbf{70.45}  & \cellcolor{brown!10}\textbf{44.97}          & \cellcolor{brown!10}\textbf{75.20}    & \cellcolor{brown!10}73.36        & \cellcolor{brown!10}66.14& \cellcolor{brown!10}\textbf{67.84}    \\

& \cellcolor{brown!10}\textbf{RUQuant{\tiny{+fine-tune}}}     & \cellcolor{brown!10}{76.12}          & \cellcolor{brown!10}69.53   & \cellcolor{brown!10}{42.92} 
& \cellcolor{brown!10}71.93 & \cellcolor{brown!10}\textbf{74.13}    & \cellcolor{brown!10}\textbf{67.01}   & \cellcolor{brown!10}66.94\\

\bottomrule
\end{tabular}
\end{center}
\label{tab:qa-llama23-w4a4}
\end{table*}

\begin{table*}[!t]
\caption{Zero-shot QA ($\uparrow$) results of LLaMA2 and LLaMA3 models under 6-bit weight-activation quantization.
}
\begin{center}
\tabcolsep=0.32cm
\renewcommand{\arraystretch}{1}
\begin{tabular}{c|l|ccccccc}
\toprule
\textbf{Model}     & \textbf{Method}  & \textbf{PIQA}  & \textbf{ARC-E} & \textbf{ARC-C} & \textbf{BoolQ} & \textbf{HellaSwag} & \textbf{WinoGrande} & \textbf{Avg.}   \\ 

\midrule
\multirow{6}{*}{\begin{tabular}[c]{@{}c@{}}LLaMA2-7B\\ W6A6\end{tabular}} 
& FP16             & 76.88          & 53.54          & 40.53          & 71.13          & 72.96              & 67.25               & 63.72          \\ \cmidrule{2-9} 
& SmoothQuant      & 75.57         & 53.62          & 39.93          & 69.54          & 71.76              & 66.14               & 62.76                     \\
& OmniQuant        & 76.55         & 53.83          & 40.96          & 68.75          & 55.89              & 65.59               & 60.26                     \\
& QLLM             & 77.48         & 52.99          & 39.33          & -              & 71.38              & 65.98               & 61.43                     \\
& DuQuant             & 76.99          & \textbf{52.99}         & \textbf{40.87}          & \textbf{70.40}         & 72.49              &67.32               & 63.51          \\

& \cellcolor{brown!10}\textbf{RUQuant}           & \cellcolor{brown!10}\textbf{77.04}   & \cellcolor{brown!10}52.97          &\cellcolor{brown!10}40.61 
& \cellcolor{brown!10}\textbf{70.40}  & \cellcolor{brown!10}\textbf{72.71}   &\cellcolor{brown!10}\textbf{67.88}   &\cellcolor{brown!10}\textbf{63.60} \\
\midrule
\multirow{6}{*}{\begin{tabular}[c]{@{}c@{}}LLaMA2-13B\\ W6A6\end{tabular}} 
& FP16             & 79.05          & 57.91          & 44.20          & 69.02          & 76.60              & 69.69               & 66.08          \\ \cmidrule{2-9} 
& SmoothQuant      & 78.29         & 57.41          & 43.86          & 69.50          & 75.02              & 66.93               & 65.17                     \\
& OmniQuant        & 78.24         & 57.58          & 43.86          & 71.10          & 75.52              & 68.35               & 65.78                     \\
& QLLM             & 78.78         & 58.29          & 43.77          & -              & 75.10              & 68.43               & 64.87                    \\
& DuQuant             & 78.62          & 56.94       & 43.43          & \textbf{68.35}          & 76.19             & 69.22              & 65.46          \\

& \cellcolor{brown!10}\textbf{RUQuant}     & \cellcolor{brown!10}\textbf{79.16}   & \cellcolor{brown!10}\textbf{57.74}  & \cellcolor{brown!10}\textbf{44.45}         & \cellcolor{brown!10}67.83    & \cellcolor{brown!10}\textbf{76.45}        & \cellcolor{brown!10}\textbf{69.77}      & \cellcolor{brown!10}\textbf{65.90}    \\
\midrule
\multirow{5}{*}{\begin{tabular}[c]{@{}c@{}}LLaMA3-8B\\ W6A6\end{tabular}} 
& FP16             & 80.85          & 77.78          & 53.41          & 81.28          & 79.16              & 72.84              & 74.22         \\ \cmidrule{2-9} 
& SmoothQuant      &78.94          & 75.88         & 49.49          &77.58          & 77.39              &70.80               & 71.68         \\
& OmniQuant        & 78.90          & 73.95          & 47.35          & 74.95          & 76.77              &70.56               & 70.41          \\
& DuQuant             & 80.20          & 77.27          & 52.05          & 80.12         & \textbf{79.14}              & 72.77               & 73.59       \\
& \cellcolor{brown!10}\textbf{RUQuant}     & \cellcolor{brown!10}\textbf{80.25}   & \cellcolor{brown!10}\textbf{77.44}  & \cellcolor{brown!10}\textbf{52.73}          & \cellcolor{brown!10}\textbf{80.40}    & \cellcolor{brown!10}78.99        & \cellcolor{brown!10}\textbf{73.24}      & \cellcolor{brown!10}\textbf{73.84}    \\
\bottomrule
\end{tabular}
\end{center}
\label{tab:qa-llama23-w6a6}
\end{table*}

\section{Experiment}
\subsection{Experimental Settings} To validate the superiority of our proposed method, we conduct two types of experiments: perplexity experiments and accuracy experiments. Following DuQuant \cite{Duquant}, during the experimental procedure, each model does not see any task-specific data, and the accuracy experiments are also conducted in zero-shot setting.

\textbf{Models and Datasets.} For perplexity experiments, we evaluate several different models by quantizing the pre-trained LLMs: LLaMA1 (7B--30B) \cite{touvron2023llama}, LLaMA2 (7B--13B) \cite{touvron2023llama2}, and LLaMA3 (8B), on the C4 \cite{raffel2020exploring} and WikiText2 \cite{merity2016pointer} datasets. For zero-shot experiments, we measure the above-mentioned models on PIQA \cite{tata2003piqa}, ARC \cite{clark2018think}, HellaSwag \cite{HellaSwag}, BoolQ \cite{clark2019boolq}, and WinoGrande \cite{sakaguchi2021winogrande} datasets. For the calibration dataset $\mathbf{X}$, we use 128 random selections of 2048-token segments from the WikiText2 dataset \cite{merity2016pointer}.

\textbf{Baselines.} We compare our method with several strong weight and activation quantization methods. (1) {Smoothquant} \cite{Smoothquant}, (2) {Omniquant} \cite{OMniquant}, (3) {QLLM} \cite{QLLM}, (4) {DuQuant} \cite{Duquant}.

 \subsection{Implementation Details}
Our main quantization levels are W4A4 and W6A6. We quantize all intermediate activations, with the exception of the Softmax output. Additionally, we apply a smooth pre-processing step to the model prior to quantization \cite{Smoothquant}, which does not introduce any additional storage or inference latency.

We observe that nearly lossless compression can be achieved even without fine-tuning by Step 2. Therefore, we propose two schemes: RUQuant and RUQuant{\scriptsize{}+fine-tune}. For RUQuant, we follow the two-step process, but do not fine-tune $\boldsymbol{\theta}$ in Step 2, which enables us to complete the quantization of a 13B parameter model in approximately one minute. For RUQuant{\scriptsize{}+fine-tune}, we use gradient descent to fine-tune $\boldsymbol{\theta}$, which improves the accuracy but significantly increases the time cost. We only present RUQuant{\scriptsize{}+fine-tune} results under the W4A4 setting, as the performance without fine-tuning is already high for W6A6, making fine-tuning unnecessary in that case. In terms of quantization method, we adopted RTN quantization and clip the weights and activations. All quantization experiments are conducted on NVIDIA L20 GPUs. For more details and specific hyperparameters, please refer to Appendix \ref{appE}.

\subsection{Analysis of Perplexity Results}
Table~\ref{tab:ppl-wikitext2} reports perplexity results of various quantization methods on LLaMA1, LLaMA2, and LLaMA3 using the WikiText2 dataset. The proposed RUQuant method consistently outperforms strong baselines like SmoothQuant, OmniQuant, QLLM, and DuQuant under both W4A4 and W6A6 settings, often matching full-precision performance. 
Overall, RUQuant provides effective compression with minimal perplexity degradation across model scales and generations, demonstrating strong robustness and generalizability. On the C4 dataset, RUQuant still demonstrate superior performance, please refer to the Appendix \ref{appH2} for detailed results.

\subsection{Analysis of Zero-Shot Results}
Tables~\ref{tab:qa-llama23-w4a4} and~\ref{tab:qa-llama23-w6a6} report zero-shot QA results on LLaMA2-7B, LLaMA2-13B, and LLaMA3-8B under 4-bit and 6-bit quantization. RUQuant consistently outperforms strong baselines such as SmoothQuant, OmniQuant, QLLM, and DuQuant.
Under 4-bit quantization, RUQuant shows clear accuracy improvements over all baselines across all models. In the 6-bit setting, it remains competitive and closely matches full-precision performance.
Overall, RUQuant demonstrates robust and generalizable performance across different models and quantization levels, particularly excelling in the 4-bit scenarios.
\subsection{Ablation Study}

\subsubsection{Influence of Different Components.} We evaluate the impact of three key components on compression performance: (1) removing the Householder reflection step, (2) removing the Givens rotation step, and (3) removing the entire Learnable Householder matrix (LH). Table \ref{tab:ablation1} shows the results under 4-bit weight-activation quantization on LLaMA2-7B and LLaMA2-13B.
As observed, removing the Householder reflection  step (w/o Householder) leads to the most significant performance degradation. In this setting, the model nearly loses its representational capacity, and fails to converge (NaN) on LLaMA2-13B. In contrast, removing the Givens rotation (w/o Givens) also causes a performance drop, but to a lesser extent. When the entire Learnable Householder matrix is removed (w/o LH), performance slightly decreases but remains close to that of the full method (RUQuant), indicating that LH can be used in precision-sensitive scenarios.

\subsubsection{Runtime.} Table \ref{tab:runtime} presents the runtime required to compress the LLaMA models of sizes 7B, 13B, and 30B. As shown, RUQuant takes only around 68 seconds to compress the 13B model, demonstrating the ability to complete compression within approximately one minute. This makes RUQuant competitive with DuQuant in terms of runtime performance. Furthermore, RUQuant shows consistent performance across all model scales, indicating good scalability and strong potential for practical deployment.

\begin{table}[!t]
\caption{Influence of different components in RUQuant  under 4-bit weight-activation quantization.}
\begin{center}
\resizebox{0.99\linewidth}{!}{
\begin{tabular}{l|cc|cc}
\toprule
\multirow{2}{*}{\textbf{Method}} & \multicolumn{2}{c|}{\textbf{LLaMA2-7B}} & \multicolumn{2}{c}{\textbf{LLaMA2-13B}} \\
 & \textbf{WikiText2} $\downarrow$ & \textbf{C4} $\downarrow$ & \textbf{WikiText2} $\downarrow$ & \textbf{C4} $\downarrow$ \\
\midrule
 w/o Householder & 33.04e3 & 27.34e3 & NaN & NaN \\
w/o Givens      & 27.42  & 35.20   & 1.32e3  & 1.54e3  \\
 w/o LH    & {6.18}  & 7.64    & 5.36 & 6.88 \\
RUQuant      & \textbf{6.17}  & \textbf{7.63}    & \textbf{5.35} & \textbf{6.87} \\
\bottomrule
\end{tabular}
}
\end{center}
\label{tab:ablation1}
\end{table}

\begin{table}[t]
\caption{Running time (in seconds) of RUQuant and DuQuant on LLaMA1 models with W4A4 quantization on WikiText2.}
\label{tab:runtime}
\centering
\begin{tabular}{c|c|c|c}
\toprule
\textbf{Method} & \textbf{LLaMA1-7B}  & \textbf{LLaMA1-13B}  & \textbf{LLaMA1-30B}  \\
\midrule
DuQuant & 44     & 66       & 136      \\
\midrule
RUQuant    & 47     & 68       & 145      \\
\bottomrule
\end{tabular}
\end{table}

\subsection{Inference Speedup}
To assess the layer-wise inference speedup enabled by our RUQuant method, we follow the same evaluation protocol and use the W4A4 kernel as described in~\cite{quarot}.
Table~\ref{tab:speedup} reports the speedup in the prefill stage for {LLaMA2-7B} and {LLaMA2-13B}, measured on a single {NVIDIA RTX 3090 GPU} with a prefill sequence length of 2048. Across all batch sizes, RUQuant consistently delivers competitive acceleration, achieving up to {1.935$\times$} speedup for {LLaMA2-7B} and {2.104$\times$} for {LLaMA2-13B}.
Notably, if we allow for a slight trade-off in performance by removing the Learnable Householder (w/o LH), the perplexity of the model increases by only {0.01} (as shown in Table~\ref{tab:ablation1}), while still outperforming {DuQuant} in accuracy. Moreover, the inference speedup becomes {identical to that of DuQuant}, making this configuration highly practical in latency-sensitive scenarios.
\begin{table}[t]
\caption{Layer-wise speedup during pre-filling stage for W4A4 quantization.}
\label{tab:speedup}
\centering
\begin{tabular}{cccc}
\toprule
\textbf{Model} & \textbf{Batch Size} & \textbf{RUQuant} &\textbf{ w/o LH}\\

\midrule
\multirow{3}{*}{\begin{tabular}[c]{c} LLaMA2-7B\\ \end{tabular}} & 1 &  $1.816\times$  &  $1.934\times$\\
&4 &  $1.845\times$ &  $1.993\times$\\
&16 &  $1.935\times$ &  $2.104\times$ \\
\midrule
\multirow{3}{*}{\begin{tabular}[c]{c} LLaMA2-13B\\ \end{tabular}} & 1 &  $1.934\times$ &  $2.154\times$\\
&4 &  $1.993\times$ &  $2.293\times$\\
&16 &  $2.104\times$ &  $2.474\times$\\
\bottomrule
\end{tabular}
\end{table}
\subsection{Memory Consumption}
We evaluate the peak memory usage of RUQuant under the same settings as DuQuant, using a W4A4 kernel on LLaMA2-7B and a single NVIDIA RTX 3090 GPU. We process 2048 tokens during the pre-filling stage and run 128 decoding steps. Results are shown in Table~\ref{tab:memery}. Compared to the FP16 baseline, RUQuant achieves up to 3.3× and 3.5× memory reduction for pre-filling and decoding respectively at batch size 1, and 2.9× and 3.6× at batch size 4. RUQuant matches DuQuant in both pre-filling and decoding stages across all batch sizes, demonstrating equivalent memory efficiency. 
\begin{table}[t]
\caption{Peak memory usage of the proposed method compared with DuQuant and FP16 during the pre-filling and decoding stages across varying batch sizes.}
\begin{center}
\begin{tabular}{c|c|cc}
\toprule
\textbf{Batch Size}        & \multicolumn{1}{c|}{\textbf{Method}}  & \textbf{Pre-filling(GB) } &  \textbf{Decoding(GB)}  \\ 
\midrule
\multirow{3}{*}{1} 
 & FP16 & 14.029  & 13.702 \\
& DuQuant & 4.287  & 3.922 \\
 & RUQuant  & 4.287     & 3.922 \\           
\midrule
\multirow{3}{*}{4}
& FP16 & 18.212  &16.939 \\
 & DuQuant & 6.228  & 4.768 \\
 & RUQuant  & 6.228     & 4.768  \\                 
\bottomrule
\end{tabular}
\end{center}
\label{tab:memery}
\end{table}

\section{Conclusion}
We propose RUQuant, a theoretically grounded, two-stage orthogonal transformation method for post-training activation quantization in large language models. By aligning non-uniform activation distributions with uniform quantizers through structured orthogonal mappings, RUQuant effectively minimizes quantization error without requiring model fine-tuning. Empirical results on a 13B model show that RUQuant achieves up to 99.8\% of full-precision accuracy with W6A6 and 97\% with W4A4 quantization in approximately one minute. In future work, we will extend the theoretical framework of RUQuant to codebook-based quantization, pursuing more aggressive memory compression.

\section{Acknowledgements}
This work was supported by National Natural Science Foundation of China (No. 62206038, 62106035), the Strategic Priority Research Program of the Chinese Academy of Sciences (No. XDA0490301), Liaoning Binhai Laboratory Project (No. LBLF-2023-01), and Xiaomi Young Talents Program.
\bibliographystyle{unsrt}
\bibliography{ref}
\clearpage
\appendix
\section{The Detailed Proof Procedure of Lloyd-Max Conditions}
\label{appA}
Here we provide the proof procedure for Eq.~\eqref{eq2} and Eq.~\eqref{eq3}.  
Let \(\mathbf{x}\) be an input vector with probability density function \(p(x)\). A \(b\)-bit Lloyd-Max quantizer aims to map \(\mathbf{x}\) to \(N = 2^b\) quantization levels \(\{q_k\}_{k=0}^{N-1}\). The input space is partitioned into \(N\) intervals \(\{[a_k, a_{k+1}]\}_{k=0}^{N-1}\). The quantizer is designed to minimize the mean squared error (MSE) between the input and its quantized representation, defined as:
\begin{equation}
    E = \sum_{k=0}^{N-1} \int_{a_k}^{a_{k+1}} (x - q_k)^2 p(x) \, dx.
\end{equation}

To find the optimal quantization points \( q_k \), we take the partial derivative of \( E \) with respect to \( q_k \) and set it to zero:
\begin{equation}
\frac{\partial E}{\partial q_k} = 0.
\end{equation}
Since the error \( E \) is a sum of integrals, and only the \( k \)-th integral depends on \( q_k \), we have:
\begin{equation}
\frac{\partial E}{\partial q_k} = \frac{\partial}{\partial q_k} \int_{a_k}^{a_{k+1}} (x - q_k)^2 p(x) \, dx.
\end{equation}
Using the chain rule, we obtain:
\begin{equation}
\frac{\partial E}{\partial q_k} = \int_{a_k}^{a_{k+1}} \frac{\partial}{\partial q_k} (x - q_k)^2 p(x) \, dx = \int_{a_k}^{a_{k+1}} -2 (x - q_k) p(x) \, dx = 0.
\end{equation}
This simplifies to:
\begin{equation}
\int_{a_k}^{a_{k+1}} x p(x) \, dx = q_k \int_{a_k}^{a_{k+1}} p(x) \, dx.
\end{equation}
Thus, the optimal quantization point \( q_k \) is the centroid of the interval \([a_k, a_{k+1}]\):
\begin{equation}
q_k = \frac{\int_{a_k}^{a_{k+1}} x p(x) \, dx}{\int_{a_k}^{a_{k+1}} p(x) \, dx}.
\end{equation}

Next, we minimize \( E \) with respect to the decision boundaries \( a_k \). The partial derivative of \( E \) with respect to \( a_k \) is:
\begin{equation}
\frac{\partial E}{\partial a_k} = 0.
\end{equation}
Since \( a_k \) affects both the \( (k-1) \)-th and \( k \)-th integrals, we have:
\begin{equation}
\frac{\partial E}{\partial a_k} = \frac{\partial}{\partial a_k} \left( \int_{a_{k-1}}^{a_k} (x - q_{k-1})^2 p(x) \, dx + \int_{a_k}^{a_{k+1}} (x - q_k)^2 p(x) \, dx \right).
\end{equation}
Applying the Leibniz integral rule, the derivative is evaluated at \( a_k \):
\begin{equation}
\frac{\partial E}{\partial a_k} = (a_k - q_{k-1})^2 p(a_k) - (a_k - q_k)^2 p(a_k) = 0.
\end{equation}
This implies:
\begin{equation}
|a_k - q_{k-1}| = |a_k - q_k|.
\end{equation}
Thus, the optimal decision boundary \( a_k \) is the midpoint between the adjacent quantization points:
\begin{equation}
a_k = \frac{q_{k-1} + q_k}{2}.
\end{equation}
In summary, the Lloyd-Max conditions for optimal quantization are:
\begin{equation}
q_k = \frac{\int_{a_k}^{a_{k+1}} x p(x) \, dx}{\int_{a_k}^{a_{k+1}} p(x) \, dx}.
\end{equation}
and
\begin{equation}
a_k = \frac{q_{k-1} + q_k}{2}.
\end{equation}

\begin{figure*}[!h]
    \label{fig:all-w}
    \centering
    \subfigure[Original]{
        \label{fig:all-w:a}
        \includegraphics[width=0.48\columnwidth]{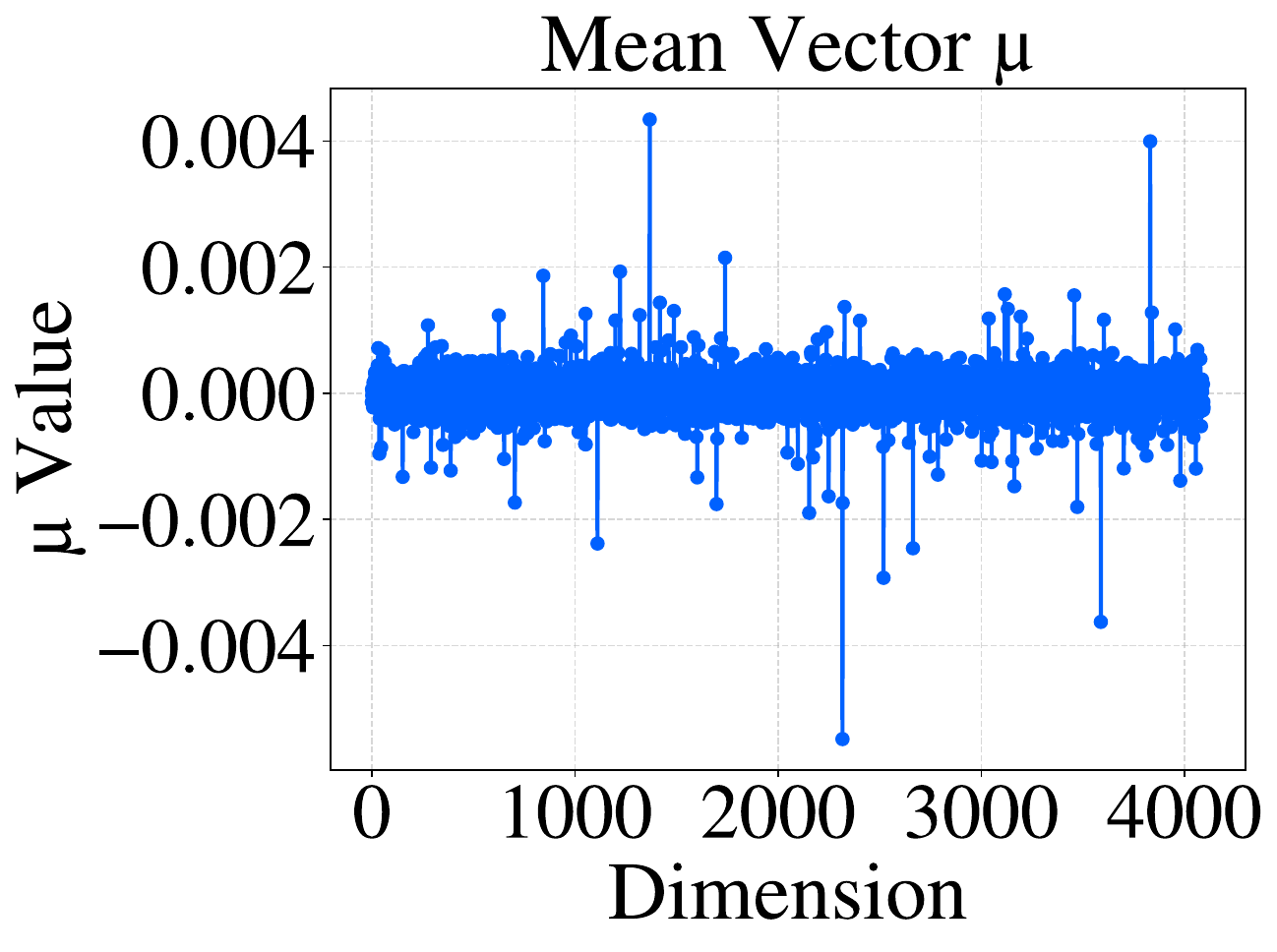}
    }
    \subfigure[RUQuant processed]{
    \label{fig:all-w:b}
    \includegraphics[width=0.48\columnwidth]{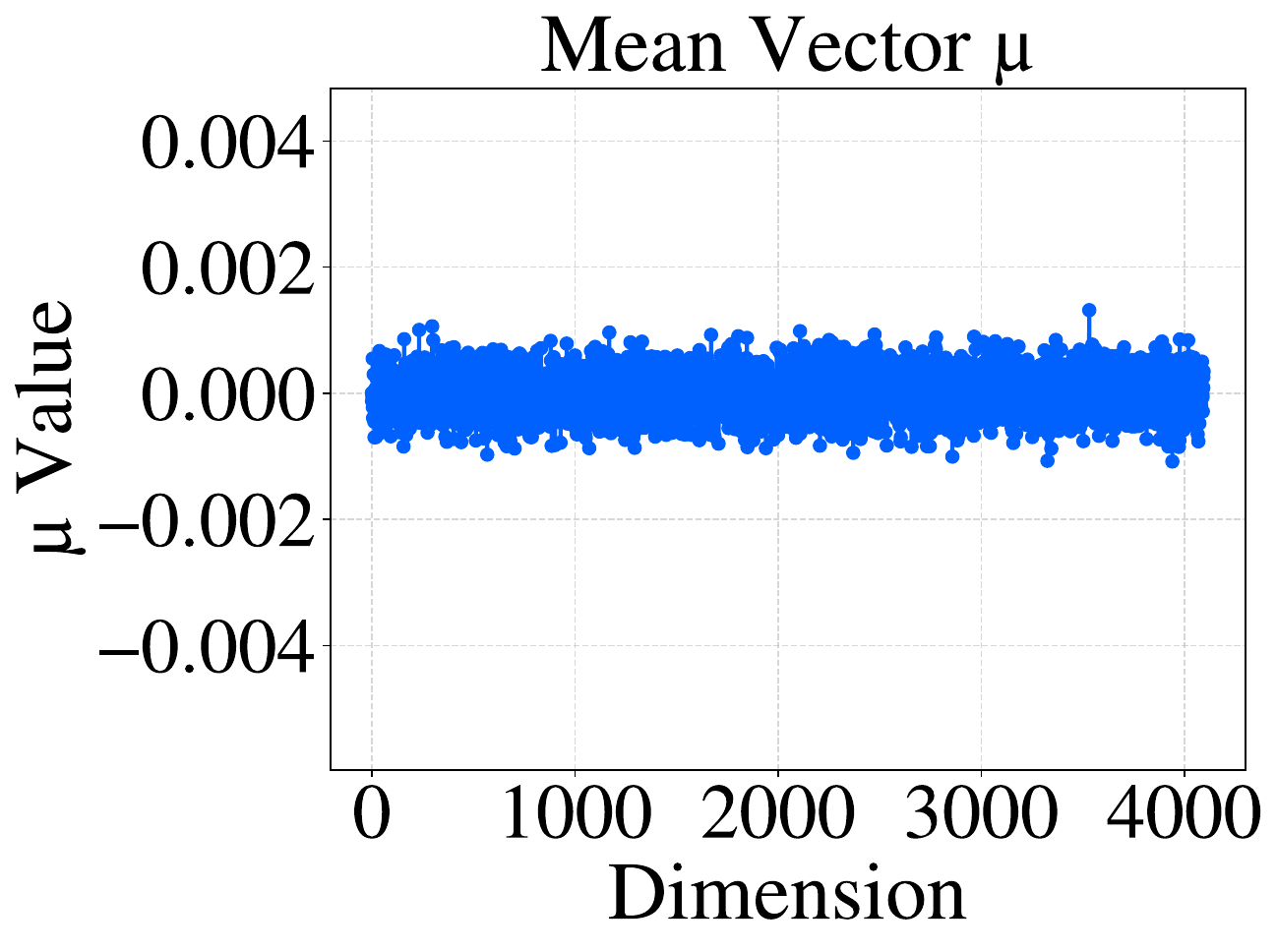}
    }
    \subfigure[Original]{
        \label{fig:all-w:c}
        \includegraphics[width=0.49\columnwidth]{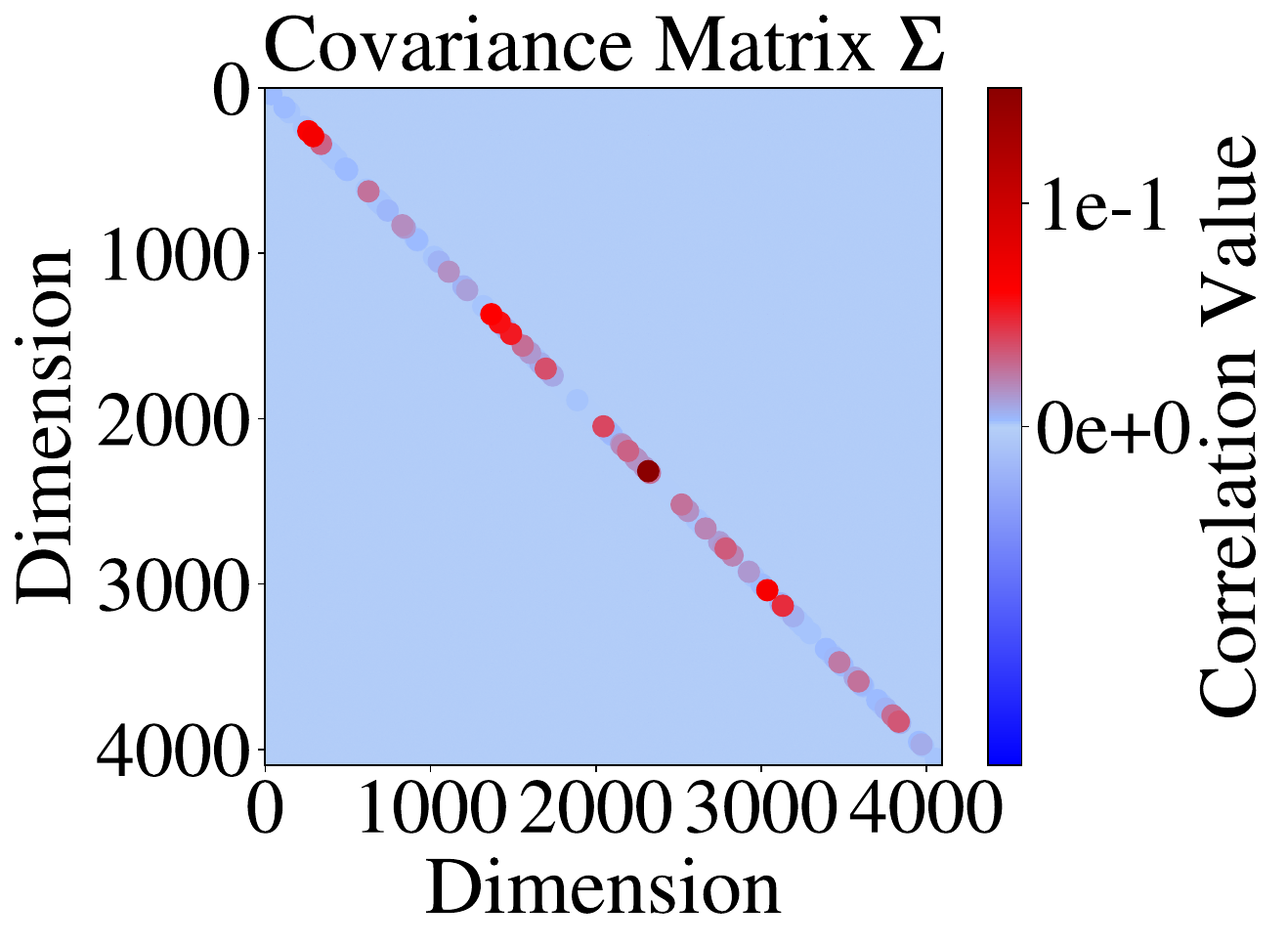}
    }
    \subfigure[RUQuant processed]{
        \label{fig:all-w:d}
        \includegraphics[width=0.49\columnwidth]{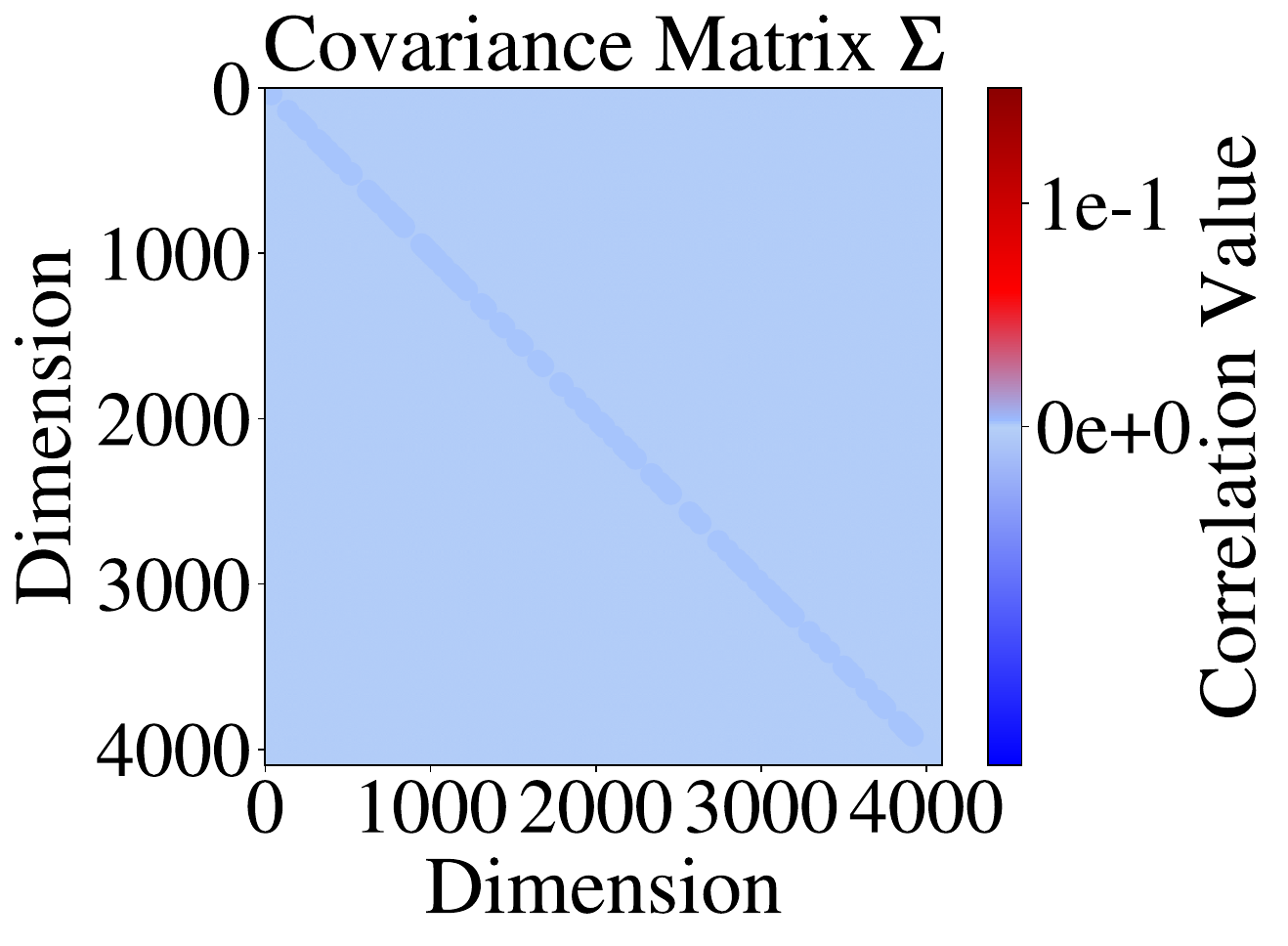}
    }
    \caption{The mean and covariance visualizations of weights before and after RUQuant processing. (a) Original mean vector. (b) Mean vector after RUQuant processing. (c) Original covariance matrix. (d) Covariance matrix after RUQuant processing.}
    \label{fig2:all-W}
\end{figure*}

\section{The Detailed Proof Procedure of Theorem \ref{theorem1}}
\label{appB}
Here we provide the proof procedure for Theorem \ref{theorem1}. We observe that the activation  \(\mathbf{X}\) follow a multivariate normal distribution \(\mathcal{N}(\boldsymbol{\mu}, \boldsymbol{\Sigma})\), where \(\boldsymbol{\mu} \in \mathbb{R}^d\) is the mean vector and \(\boldsymbol{\Sigma} \in \mathbb{R}^{d \times d}\) is the covariance matrix. Moreover, it is assumed that \(\|\boldsymbol{\mu}\|_2^2 \gg \text{tr}(\boldsymbol{\Sigma})\), which aligns with previous findings \cite{Smoothquant, AWQ} that the variance within rows of \(\mathbf{X}\) is small, while the variance between rows is large.

Given a sampled vector \(\mathbf{x} \sim \mathcal{N}(\boldsymbol{\mu}, \boldsymbol{\Sigma})\) and a uniformly distributed vector \(\hat{\mathbf{u}} \sim \mathbf{U}(-1, 1)\), we define \(\mathbf{u} = \frac{\|\mathbf{x}\|_2}{\|\hat{\mathbf{u}}\|_2} \hat{\mathbf{u}}\). We then determine a unique orthogonal matrix \(\mathbf{Q}\) (computed via Householder or Givens transformations) such that \(\mathbf{u} = \mathbf{Q} \mathbf{x}\). For another sample \(\mathbf{x}_{\text{new}} \sim \mathcal{N}(\boldsymbol{\mu}, \boldsymbol{\Sigma})\) from \(\mathbf{X}\) with the same distribution as \(\mathbf{x}\), considering the linear transformation of Gaussian vectors, we have:
\begin{equation}
\mathbf{Q} \mathbf{x}_{\text{new}} \sim \mathcal{N}(\mathbf{Q} \boldsymbol{\mu}, \mathbf{Q} \boldsymbol{\Sigma} \mathbf{Q}^T).
\end{equation}
Given any permutation matrix \(\mathbf{P}\) and diagonal sign matrix \(\mathbf{D}\):
\begin{equation}
\mathbf{P} \mathbf{D} \mathbf{u} \overset{d}{=} \mathbf{u},
\end{equation}
and considering \(\mathbf{Q} \mathbf{x} = \mathbf{u}\), we obtain:
\begin{equation}
\mathbf{P} \mathbf{D} \mathbf{Q} \mathbf{x} \overset{d}{=} \mathbf{Q} \mathbf{x}.
\end{equation}
Since \(\mathbf{x}\) is arbitrary, we have:
\begin{equation}
\mathbf{P} \mathbf{D} \mathbf{Q} \overset{d}{=} \mathbf{Q}.
\end{equation}
Thus,
\begin{equation}
\mathbb{E}_{\mathbf{Q}}[\mathbf{Q} \boldsymbol{\Sigma} \mathbf{Q}^T] = \mathbb{E}_{\mathbf{Q}}[\mathbf{P} \mathbf{D} \mathbf{Q} \boldsymbol{\Sigma} \mathbf{Q}^T \mathbf{D} \mathbf{P}^T] = \mathbf{P} \mathbf{D} \mathbb{E}_{\mathbf{Q}}[\mathbf{Q} \boldsymbol{\Sigma} \mathbf{Q}^T] \mathbf{D} \mathbf{P}^T.
\end{equation}
Considering the algebra generated by the symmetric group \(\{\mathbf{P} \mathbf{D}\}\), which is irreducible in \(\mathbb{R}^d\), any matrix commuting with all \(\mathbf{P} \mathbf{D}\) must be a scalar matrix \(c \mathbf{I}\). Taking the trace operation, we have:
\begin{equation}
\text{tr}(c \mathbf{I}) = c d = \text{tr}(\mathbb{E}_{\mathbf{Q}}[\mathbf{Q} \boldsymbol{\Sigma} \mathbf{Q}^T]) =\mathbb{E}_{\mathbf{Q}}[\text{tr}(\boldsymbol{\Sigma})] = \text{tr}(\boldsymbol{\Sigma}).
\end{equation}
Solving for \(c\), we obtain:
\begin{equation}
c = \frac{\text{tr}(\boldsymbol{\Sigma})}{d}.
\end{equation}
Thus, we have:
\begin{equation}
\mathbb{E}_{\mathbf{Q}}[\mathbf{Q} \boldsymbol{\Sigma} \mathbf{Q}^T] = \frac{\text{tr}(\boldsymbol{\Sigma})}{d} \mathbf{I}.
\end{equation}
Given that the diagonal elements of the activation covariance matrix are small, we conclude that the covariance matrix is:
\begin{equation}
\mathbb{E}_{\mathbf{Q}}[\mathbf{Q} \boldsymbol{\Sigma} \mathbf{Q}^T] = \lambda \mathbf{I}, \quad \lambda \approx 0,
\end{equation}
indicating that the transformed covariance expectation is isotropic noise.

Let \(\mathbf{x} = \boldsymbol{\mu} + \epsilon\), where \(\epsilon \sim \mathcal{N}(0, \boldsymbol{\Sigma})\). Since \(\mathbf{u} = \mathbf{Q} \mathbf{x} = \mathbf{Q} \boldsymbol{\mu} + \mathbf{Q} \epsilon\), we have:
\begin{equation}
\mathbf{Q} \boldsymbol{\mu} = \mathbf{u} - \mathbf{Q} \epsilon.
\end{equation}
Given \(\|\mathbf{x}\|_2^2 = \|\boldsymbol{\mu}\|_2^2 + 2 \boldsymbol{\mu}^T \epsilon + \|\epsilon\|_2^2\), and knowing that \(\epsilon \sim \mathcal{N}(0, \boldsymbol{\Sigma})\), \(\mathbb{E}[\boldsymbol{\mu}^T \epsilon] = 0\), and \(\mathbb{E}[\|\epsilon\|_2^2] = \text{tr}(\boldsymbol{\Sigma})\), we obtain:
\begin{equation}
\mathbb{E}[\|\mathbf{x}\|_2^2] = \|\boldsymbol{\mu}\|_2^2 + \text{tr}(\boldsymbol{\Sigma}).
\end{equation}
Since \(\|\boldsymbol{\mu}\|_2^2 \gg \text{tr}(\boldsymbol{\Sigma})\), we have:
\begin{equation}
\|\mathbf{u}\|_2^2 = \|\mathbf{x}\|_2^2 \approx \|\boldsymbol{\mu}\|_2^2 = \|\mathbf{Q} \boldsymbol{\mu}\|_2^2.
\end{equation}
Moreover,
\begin{equation}
\text{cos} \langle \mathbf{u}, \mathbf{Q} \boldsymbol{\mu} \rangle = \frac{\mathbf{u}^{\top} \cdot \mathbf{Q} \boldsymbol{\mu}}{\|\mathbf{u}\|_2 \cdot \|\mathbf{Q} \boldsymbol{\mu}\|_2} = \frac{\mathbf{u}^{\top} \cdot \mathbf{Q} \boldsymbol{\mu}}{\|\mathbf{x}\|_2 \cdot \|\boldsymbol{\mu}\|_2}.
\end{equation}
Thus,
\begin{equation}
\mathbb{E}[\text{cos} \langle \mathbf{x}_u, \mathbf{Q} \boldsymbol{\mu} \rangle] = \mathbb{E}\left[\frac{\|\boldsymbol{\mu}\|_2^2 + \boldsymbol{\mu}^{\top}  \epsilon}{\|\mathbf{x}\|_2 \cdot \|\boldsymbol{\mu}\|_2}\right] \approx 1.
\end{equation}
Therefore,
\begin{equation}
\mathbf{Q} \boldsymbol{\mu} \approx \mathbf{u}.
\end{equation}
In summary, the transformed \(\mathbf{X}\) is distributed around the uniform vector \(\mathbf{u}\), with each dimension exhibiting small isotropic fluctuations.

\section{The Detailed Proof Procedure of Theorem \ref{theorem2}}
\label{appC}
Here we provide the proof procedure for Theorem \ref{theorem2}. Owing to the equivalent transformation, the weight matrix \(\mathbf{W}\) in a linear layer of a large-scale model is subject to the following transformation:
\begin{equation}
\mathbf{W} \leftarrow \mathbf{W} \mathbf{Q}^{\top}.
\end{equation}
The row vectors \(\mathbf{w}^{\top}\) of \(\mathbf{W}\) follow a multivariate normal distribution \cite{Post4}  \(\mathcal{N}(\boldsymbol{\mu}_W, \boldsymbol{\Sigma}_W)\), where \(\boldsymbol{\mu}_W \in \mathbb{R}^d\) is the mean vector and \(\boldsymbol{\Sigma}_W \in \mathbb{R}^{d \times d}\) is the covariance matrix, we observe that the distribution of \(\mathbf{W}\) is relatively uniform, with the mean vector \(\boldsymbol{\mu}_W \approx \mathbf{0}\). Therefore:
\begin{equation}
\mathbf{Q} \mathbf{w} \sim \mathcal{N}(\mathbf{0}, \mathbf{Q} \boldsymbol{\Sigma}_W \mathbf{Q}^T).
\end{equation}

The covariance matrix after transformation still satisfies:
\begin{equation}
\mathbb{E}_{\mathbf{Q}}[\mathbf{Q} \boldsymbol{\Sigma}_W \mathbf{Q}^T] = \frac{\text{tr}(\boldsymbol{\Sigma}_W)}{d} \mathbf{I}.
\end{equation}
Thus, the covariance dominates the shape of the distribution, and the expected covariance is isotropic noise. Consequently, the transformed \(\mathbf{w}^{\top}\) is expected to be distributed isotropically, and \(\mathbf{Q}\) also serves to smooth \(\mathbf{W}\).

\section{Visual Analysis of $\mathbf{W}$ and $\mathbf{WQ}_c^\top$}
\label{appD}
Figure \ref{fig2:all-W} shows the effect of the rotation matrix on $\mathbf{W}$. It can be found that the rotation matrix solved according to the characteristics of $\mathbf{X}$ can also have a smoothing effect on $\mathbf{W}$.

\begin{algorithm}[!t]
\caption{RUQuant quantization algorithm}
\label{algorithm:\QQQ}
\parbox[t]{\linewidth}{\textbf{Input}: Transfomer Block $\mathcal{T}$, Block Size $B$, Epoch $\mathcal{E}$ , Number of Random Permutations $\lambda$, Number of Sampling Times $K$, Zigzag Permutation Times $T$.}\\
\parbox[t]{\linewidth}{\textbf{Output}: $\mathcal{Q}(\mathbf{W}), \mathcal{Q}(\mathbf{X})$}\\
\begin{algorithmic}[1]
\FOR{$\text{layer in} \ \mathcal{T}  $}
    \STATE $\mathbf{X} \gets \text{layer.input}$ 
    \STATE $\mathbf{W}\gets\text{layer.weight}$ 
    \FOR{$i =1 \ \text{to} \ T$}
        \STATE $\mathbf{X},\mathbf{W}\gets\text{Rotation}(\mathbf{X},\mathbf{W},B,K,\lambda)$
        \STATE $\mathbf{X}, \mathbf{Z}\gets$\text{Zigzag\_Permutation}($\mathbf{X}$)
        \STATE $\mathbf{W}\gets\mathbf{WZ}^\top$
    \ENDFOR
    \STATE $\mathbf{X},\mathbf{W}\gets\text{Rotation}(\mathbf{X},\mathbf{W},B,K,\lambda)$
    \STATE $\mathbf{x}\gets\text{Sample}(\mathbf{X})$
    \STATE $\mathbf{u}\gets\mathcal{U}(-1,1)^n$
    \STATE $\mathbf{u}\gets \mathbf{u}(\|\mathbf{x}\|/\|\mathbf{u}\|)$
    \STATE $\boldsymbol{\theta}\gets\mathbf{x}-\mathbf{u}$
    \STATE $\mathbf{X}\gets \mathbf{X}-(2/\boldsymbol{\theta}^\top\boldsymbol{\theta}) \boldsymbol{\theta}(\boldsymbol{\theta}^\top\mathbf{X})$
    \STATE $\mathbf{W}\gets\mathbf{W}-(2/\boldsymbol{\theta}^\top\boldsymbol{\theta}) \mathbf{W} \boldsymbol{\theta} \boldsymbol{\theta}^\top$
\ENDFOR
\FOR{$j =1 \ \text{to} \ \mathcal{E}$}
    \STATE $L\gets \text{objective function from Eq. \eqref{eq8}} $
    \STATE $\boldsymbol{\theta} \gets \text{adam}(\boldsymbol{\theta},\frac{\partial L}{\partial \boldsymbol{\theta}})$
\ENDFOR
\STATE $\mathbf{W}\gets\mathcal{Q}(\mathbf{W})$
\STATE $\mathbf{X}\gets\mathcal{Q}(\mathbf{X})$
\end{algorithmic}
\end{algorithm}

\begin{algorithm}[!h]
\caption{Rotation}
\label{algorithm:Rotation}
\parbox[t]{\linewidth}{\texttt{func Rotation($\mathbf{X},\mathbf{W},B,K,\lambda$)}}\\
\begin{algorithmic}[1]
\STATE shapeX $\gets \mathbf{X}$.shape
\STATE shapeW $\gets \mathbf{W}$.shape
\STATE $\mathbf{X} \gets \mathbf{X}$.reshape(-1,B)
\STATE $\mathbf{W} \gets \mathbf{W}$.reshape(B,-1)
\STATE $\mathbf{Q}\gets \mathbf{I}_{B\times B} $
\FOR{$k = 1$ to $K$}
    \STATE $\mathbf{X} \gets $Sample($\mathbf{X}$)
    \STATE $\mathbf{H} \gets$ Aig \ref{algorithm1}($\mathbf{x}$)
    \STATE $\mathbf{X} \gets \mathbf{HX}$
    \STATE $\mathbf{Q} \gets \mathbf{HQ}$
    \STATE $\mathbf{X} \gets $Sample($\mathbf{X}$)
    \STATE $\mathbf{R} \gets$ Aig \ref{algorithm2}($\mathbf{x},\lambda$)
    \STATE $\mathbf{X} \gets \mathbf{RX}$
    \STATE $\mathbf{Q} \gets \mathbf{GQ}$
\ENDFOR
\STATE $\mathbf{X}\gets\mathbf{X}.\text{reshape(shapeX)}$
\STATE $\mathbf{W}\gets\mathbf{WQ}^\top.\text{reshape(shapeW)}$
\RETURN $\mathbf{W}$, $\mathbf{X}$
\end{algorithmic}
\end{algorithm}

\begin{table*}[!t]
\caption{The perplexity results ($\downarrow$) of the LLaMA1, LLaMA2 and LLaMA3 model on the C4 dataset.}
\begin{center}
\tabcolsep=0.45cm
\begin{tabular}{c|l|cccccccc}
\toprule
\textbf{\#Bit}        & \multicolumn{1}{c|}{\textbf{Method}} & \textbf{1-7B} & \textbf{1-13B} & \textbf{1-30B}  & \textbf{2-7B} & \textbf{2-13B} & \textbf{3-8B} \\ 
\midrule
FP16 &-                         & 6.92         & 6.49         &  5.92              & 6.84         &  6.41         & 8.62          \\
\midrule
\multirow{2}{*}{W4A4}        
                    & DuQuant                                 & 7.60          & 7.00           & 6.26                    & 7.63          & 6.89           & 11.45          \\
                      
                      & \cellcolor{brown!10}\textbf{RUQuant}           & \cellcolor{brown!10}\textbf{7.57}     
                      & \cellcolor{brown!10}\textbf{6.91}           & \cellcolor{brown!10}\textbf{6.25}
                      & \cellcolor{brown!10}\textbf{7.62}          
                      & \cellcolor{brown!10}\textbf{6.87}           & \cellcolor{brown!10}\textbf{ 11.39}           \\
                      
\bottomrule
\multirow{2}{*}{W6A6} 
                      & DuQuant                                 & 6.96          & 6.52           & 5.49                     & 6.89          & 6.44           & 8.78            \\
                     
                      & \cellcolor{brown!10}\textbf{RUQuant}          & \cellcolor{brown!10}\textbf{6.95}         & \cellcolor{brown!10}\textbf{6.51}           & \cellcolor{brown!10}\textbf{5.94}                    & \cellcolor{brown!10}\textbf{6.88}         & \cellcolor{brown!10}\textbf{6.43}           & \cellcolor{brown!10}\textbf{8.75}          \\
                     
\bottomrule
\end{tabular}
\end{center}
\label{tab:ppl-C4}
\end{table*}

\section{Implementation Details}
\label{appE}
In this work, all quantization experiments are conducted on NVIDIA L20 GPUs, which feature 40GB of memory. The sequence length is set to 2048 for all evaluation tasks.
In our rotation matrix experiments, we set the rotation block size to $B$ = 128 and the number of sampling times to $K$ = 16. Following the DuQuant method, we adopt a single zigzag permutation ($T=1$) to improve efficiency.
Before applying rotation, we perform a smoothing operation with a smoothing factor set to 0.6. During RTN quantization of the rotated weights and activations, we apply appropriate clipping: 1) For the W4A4 setting, We clip the maximum activation values in all projection blocks with a clipping ratio of 0.9 and clip the maximum values in the weight matrices with a clipping ratio of 0.8. 2) For the W6A6 setting, Since RUQuant yields better rotation performance in this case, we do not apply any clipping to weights or activations (clipping ratio set to 1).
When fine-tuning the LH component, we only update the parameter $\boldsymbol{\theta}$, while keeping all other settings (e.g., clipping ratio and smoothing factor) the same as in the non-fine-tuned configuration. The fine-tuning is conducted with the following settings: epoch = 20, batch size = 1, and learning rate = $1 \times 10^{-2}$.

\section{Algorithm}
\label{APPG}
Algorithm \ref{algorithm:\QQQ} shows the detailed process of RUQuant, and Algorithm \ref{algorithm:Rotation} shows the Rotation function. Specifically, for each Transformer block \( \mathcal{T} \), we perform the operations in two steps. In the first step, we apply the Householder and Givens Transformations  to all linear layers within the Transformer block. Initially, the input \( \mathbf{X} \) is divided into blocks of size \( B \). Subsequently, all blocks share a common rotation matrix \( \mathbf{Q}_c \), and through the composite linear transformation of Householder and Givens transforms, an exact mapping from non-uniform to uniform distribution is achieved. To reduce the variance between blocks, we further employ a zigzag permutation and a second rotation operation. In the second step, we initialize the Learnable Householder matrix parameters \( \boldsymbol{\theta} \) using Algorithm \ref{algorithm1} and fine-tune \( \boldsymbol{\theta} \) based on the changes in block outputs before and after quantization. This process aims to further eliminate the impact of block division and align the quantization loss with the model loss.

\begin{table}[t]
\caption{5-shot MMLU performance of LLaMA2-7B and 13B under W4A4 quantization.}
\label{tab:mmlu_w4a4_0shot}
\centering
\begin{tabular}{c|c|c|c|c|c}
\toprule
\textbf{Model} & \textbf{Method} & \textbf{STEM} & \textbf{Hums} & \textbf{Social} & \textbf{Others} \\
\midrule
\multirow{3}{*}{7B}
& FP16      & 36.88 & 43.25 & 51.74 & 52.41 \\
& DuQuant   & 30.62 & 34.16 & 40.53 & 42.97 \\
& \textbf{RuQuant}  & \textbf{33.10} & \textbf{37.56} & \textbf{43.74} & \textbf{46.58} \\
\midrule
\multirow{3}{*}{13B}
& FP16      & 44.23 & 54.41 & 63.47 & 60.67 \\
& DuQuant   & 39.10 & 46.48 & \textbf{57.69} & 55.64 \\
& \textbf{RuQuant}  & \textbf{39.96} & \textbf{47.37} & \textbf{57.69} & \textbf{55.89} \\
\bottomrule
\end{tabular}
\end{table}

\begin{table}[t]
\caption{0-shot MMLU performance of LLaMA2-7B and 13B under W4A4 quantization.}
\label{tab:mmlu_w4a4_5shot}
\centering
\begin{tabular}{c|c|c|c|c|c}
\toprule
\textbf{Model} & \textbf{Method} & \textbf{STEM} & \textbf{Hums} & \textbf{Social} & \textbf{Others} \\
\midrule
\multirow{3}{*}{7B}
& FP16      &33.96&39.36&47.94&45.92 \\
& DuQuant   & 29.92&30.24&36.30&35.38 \\
& \textbf{RuQuant}  & \textbf{30.88} & \textbf{32.86} & \textbf{39.65} & \textbf{36.18} \\
\midrule
\multirow{3}{*}{13B}
& FP16      & 42.35&47.76&60.71&59.47 \\
& DuQuant   & 37.64&44.97&55.15&52.28 \\
& \textbf{RuQuant}  & \textbf{39.99} & \textbf{44.42} & \textbf{55.18} & \textbf{54.19} \\
\bottomrule
\end{tabular}
\end{table}

\section{More Results}
\label{appH2}
Table \ref{tab:ppl-C4} shows the test results of RUQuant on the C4 dataset. As can be seen from the table, RUQuant still has certain advantages on the C4 dataset, and the performance of the quantized model also surpasses DuQuant. The experiments demonstrate that RUQuant consistently achieves lower perplexity across multiple model sizes and quantization levels. 

Table \ref{tab:mmlu_w4a4_0shot} and \ref{tab:mmlu_w4a4_5shot} show the MMLU results under 0-shot and 5-shot settings. It can be seen that RUQuant achieves better performance than DuQuant across domains, further demonstrating its effectiveness on challenging downstream tasks. 

\end{document}